\documentclass[runningheads]{llncs}

\usepackage{accv}

\usepackage{accvabbrv}
\usepackage{soul}

\usepackage{url}
\usepackage{booktabs}
\usepackage{amsfonts}
\usepackage{nicefrac}
\usepackage{microtype}
\usepackage{lipsum}

\usepackage{graphicx}
\usepackage{subcaption}
\usepackage{wrapfig}
\usepackage{caption}
\usepackage{float}

\usepackage{multirow}
\usepackage{threeparttable}

\usepackage{amsmath}
\usepackage{amssymb}
\usepackage{bm}
\usepackage{dsfont}

\usepackage{algorithm}
\usepackage{algorithmic}
\usepackage{listings}
\usepackage{verbatim}
\usepackage{comment}

\usepackage{pifont}

\usepackage{textcomp}
\usepackage{relsize}

\usepackage{cancel}

\usepackage{appendix}
\usepackage{afterpage}

\usepackage{etoolbox}
\makeatletter
\AfterEndEnvironment{algorithm}{\let\@algcomment\relax}
\AtEndEnvironment{algorithm}{\kern2pt\hrule\relax\vskip3pt\@algcomment}
\let\@algcomment\relax
\newcommand\algcomment[1]{\def\@algcomment{\footnotesize#1}}
\renewcommand\fs@ruled{\def\@fs@cfont{\bfseries}\let\@fs@capt\floatc@ruled
  \def\@fs@pre{\hrule height.8pt depth0pt \kern2pt}%
  \def\@fs@post{}%
  \def\@fs@mid{\kern2pt\hrule\kern2pt}%
  \let\@fs@iftopcapt\iftrue}
\makeatother

\newcommand{\model}{SNAP\xspace}
\newcommand{\std}[1]{{\fontsize{2.2}{2.2}\selectfont$_{.#1}$}}

\usepackage[table]{xcolor}
\definecolor{RealNeg}{HTML}{D72638}
\definecolor{SynNeg}{HTML}{6F2DA8}
\definecolor{RealPos}{HTML}{009B55}
\definecolor{colorarch}{gray}{0.90}
\definecolor{colormethod}{RGB}{247,224,213} 

\usepackage[accsupp]{axessibility}
\usepackage{orcidlink}
\usepackage{hyperref}

\begin{document}

\title{
What Makes Synthetic Hard Negatives Work in Vision-Language Pretraining?
}

\titlerunning{What Makes Synthetic Hard Negatives Work in VLP?}

\author{
Nikos Giakoumoglou\inst{1} \and
Paschalis Giakoumoglou\inst{2} \and
Andreas Floros\inst{1} \and \\
Kleanthis Marios Papadopoulos\inst{1} \and
Tania Stathaki\inst{1}
}

\authorrunning{N.~Giakoumoglou et al.}

\institute{
Imperial College London, London, UK \and
Centre for Research and Technology Hellas, Thessaloniki, Greece\\
\email{nikos@imperial.ac.uk}\\
}

\maketitle


\begin{abstract}
Synthetic hard negatives generated in the representation space have proven effective for unimodal self-supervised learning, but transferring this idea to vision-language pretraining is not straightforward. We analyze six representation-space synthesis strategies and identify two failure modes in their transfer to vision-language pretraining: cross-modal constructions that produce overly easy negatives or pull them toward the query, and intra-modal constructions that incorporate the matched positive. We also observe logit-scale saturation when training with synthetic hard negatives and a learnable temperature, and find that fixing the temperature improves downstream performance. Using this geometric analysis we propose \model, which generates intra-modal hard negatives that never involve the positive from either modality, avoiding both failure modes entirely. \model is model-agnostic, requires no external generative models, and adds less than 10\% training time overhead. Evaluated on top of CLIP and FLIP across multiple architectures and datasets, \model delivers consistent improvements on zero-shot retrieval, zero-shot classification, and linear probe evaluation.


\end{abstract}

\section{Introduction}
\label{sec:introduction}

Contrastive pretraining using weakly-supervised image-text pairs collected from the web has become the dominant paradigm for learning generic visual representations, progressively replacing supervised pretraining on large labeled datasets~\cite{imagenet}. The core idea is to jointly learn an aligned embedding space for images and text from paired data, maximizing the similarity of matched (``\textit{positive}'') image-text pairs while pushing apart mismatched (``\textit{negative}'') pairs via a batch-level InfoNCE loss~\cite{oord2019cpc,hadsell2006dimensionality}. Seminal works CLIP~\cite{radford2021clip}, ALIGN~\cite{jia2021align}, and BASIC~\cite{pham2023basic} demonstrated remarkable zero-shot transfer capabilities by training on hundreds of millions of image-text pairs. However, standard CLIP-style training samples image--text pairs randomly,
without explicitly controlling the difficulty or multiplicity of in-batch
negatives. Although higher-similarity negatives already receive greater
softmax weight in InfoNCE, their availability depends on batch composition.
Hard-negative sampling and synthesis provide explicit control over this
training signal~\cite{kalantidis2020mochi,robinson2020contrastive}.

In unimodal self-supervised learning, synthetic hard negatives generated directly in the representation space have proven highly effective~\cite{kalantidis2020mochi,giakoumoglou2024synco,giakoumoglou2025unsupervised,giakoumoglou2025fake}, offering a fast and lightweight alternative to input-space augmentation. This raises a natural question: \textit{can representation-space synthetic negatives transfer to vision-language pretraining?} Prior work has explored hard negatives for vision-language models primarily in the \textbf{input space}: rule-based text perturbations~\cite{yuksekgonul2023negclip,zhang2023contrasting}, LLM-based caption rewrites~\cite{fan2023laclip,patel2024tripletclip}, or diffusion-generated negative images~\cite{patel2024tripletclip}. These approaches incur significant computational overhead and typically address only one modality at a time (\cf \cref{sec:related}). Representation-space generation would sidestep all of these issues. However, as we show in this paper, the transfer is \textbf{not straightforward}, and understanding \textit{why} it fails reveals fundamental constraints imposed by multimodal representation geometry.

Unlike the single-modality setting where all representations inhabit one homogeneous space, vision-language models align two distinct modalities within a shared embedding space. This dual-domain structure introduces two geometric failure modes with no unimodal analogue. First, the \textit{cross-modal} synthetic negatives that were tested, produced by blending image and text embeddings, fall into the modality gap~\cite{liang2022mind}: a void between the image and text cones that real embeddings never occupy, making such samples trivially separable and uninformative. Second, restricting synthesis to a single modality but involving the positive pair causes \textit{positive leakage}: the synthetic ``negative'' inherits genuine positive signal, sending contradictory gradients that confuse optimization. We characterize these geometric effects using representation-level
diagnostics. Separately, we observe logit-scale saturation when synthetic
hard negatives are introduced with a learnable temperature.
Our temperature ablation supports combining positive-free intra-modal
synthesis with a fixed temperature
(\cref{sec:geometry,sec:temperature-divergence}). 

We make the following \textbf{contributions}:
\textbf{(i)} We identify and characterize two geometric failure modes that prevent naive transfer of representation-space synthetic negatives to vision-language pretraining, \ie, modality gap collapse and positive leakage, and provide quantitative diagnostics for each (\cref{sec:failure-modes}).
\textbf{(ii)} Guided by this analysis, we propose \model, a model-agnostic method that generates synthetic hard negatives directly in the representation space for both image and text modalities, applicable on top of InfoNCE-based frameworks such as CLIP and FLIP (\cref{sec:snap-method}).
\textbf{(iii)} We conduct comprehensive evaluations across multiple architectures (ViT-B/16, ViT-B/32, RN-50), datasets (CC3M, CC12M), and downstream tasks, \eg, zero-shot retrieval (\cref{tab:retrieval}), zero-shot classification (\cref{tab:zeroshot}), linear probe evaluation (\cref{tab:linear}), demonstrating broad improvements across the evaluated settings.

\section{Related Work}
\label{sec:related}

\subsection{Vision-Language Contrastive Pretraining}
\label{sec:related-vlp}

CLIP~\cite{radford2021clip} and ALIGN~\cite{jia2021align} established the standard approach for vision-language contrastive pretraining: learning aligned image-text representations via a batch-level InfoNCE loss~\cite{oord2019cpc} over large-scale web-crawled data. LiT~\cite{zhai2022lit} showed that freezing a pretrained vision encoder and training only the text tower can be surprisingly effective, while SigLIP~\cite{zhai2023siglip} replaced the softmax-based InfoNCE with a pairwise sigmoid formulation that removes the need for global normalization. BASIC~\cite{pham2023basic} demonstrated the benefits of jointly scaling both encoders and data. Since then, a wide range of methods has advanced the field through modifications to the contrastive objective~\cite{yao2022filip,yang2022unicl,goel2022cyclip,lavoie2024llip}, integration of self-supervised objectives~\cite{mu2022slip,li2022declip,naeem2023silc,zhou2023nclip,maninis2025tips,singh2022flava}, architectural changes~\cite{cao2025flame,ranasinghe2023clippy,yu2022coca,tschannen2025siglip2}, training efficiency improvements~\cite{li2023flip,yang2023alip}, scaling~\cite{sun2023evaclip,sun2024evaclip18b,yuan2021florence,xiao2023florence2,open_clip}, and improved training signals via synthetic captions~\cite{zhang2024longclip,zheng2024dreamlip,fan2023laclip,wu2024lotlip,liu2024clips}. \model is complementary to all these directions, augmenting the InfoNCE loss with representation-space synthetic negatives without modifying the architecture, data pipeline, or base objective.

\subsection{Hard Negative Sampling}
\label{sec:related-negatives}

The quality and difficulty of negative samples are critical for contrastive learning~\cite{schroff2015facenet,sohn2016npairs,wu2017sampling,giakoumoglou2024review}. In \textbf{vision}, synthetic hard negatives generated in the representation space have been shown to significantly improve self-supervised representation learning~\cite{kalantidis2020mochi,robinson2020contrastive,giakoumoglou2024synco,giakoumoglou2025unsupervised,giakoumoglou2025syncov2}. MoCHi~\cite{kalantidis2020mochi} mixes hard-negative
features, while SynCo~\cite{giakoumoglou2024synco} studies six synthesis
strategies that underpin our analysis.
In \textbf{vision-language learning}, linguistic
perturbations~\cite{yuksekgonul2023negclip,zhang2023contrasting}
construct negative captions, and DiHT~\cite{radenovic2023diht} increases the contribution
of hard negatives through importance sampling while NegCLIP~\cite{yuksekgonul2023negclip} and related methods~\cite{zhang2023contrasting} construct negatives through linguistic perturbations. In representation space, $m^2$-Mix within $m^3$-Mix~\cite{oh2023geodesic} geodesically mixes another matched image--text pair to form cross-modal negatives. \model instead synthesizes query-specific negatives within each modality, using hard-negative mixing and noise during pretraining from scratch.

\subsection{Synthetic Augmentation for Vision-Language Models}
\label{sec:related-synthetic}

Several recent methods synthesize additional training data for CLIP-style pretraining, but these approaches operate in the \textbf{input space} and rely on large external foundation models. On the text side, captions are rewritten or generated using LLMs~\cite{fan2023laclip,patel2024tripletclip}, captioning models~\cite{yang2023alip,liu2024clips}, or multimodal LLMs~\cite{zheng2024dreamlip,wu2024lotlip,lai2024veclip,liu2024mllmsaugmented}. On the image side, TripletCLIP~\cite{patel2024tripletclip} generates hard negative images using text-to-image diffusion models (\eg, SDXL-Turbo~\cite{podell2023sdxl}) conditioned on LLM-generated captions. These methods introduce external knowledge from models that may overlap with downstream benchmarks, making it difficult to disentangle representational improvements from data leakage, and add substantial computational overhead. \model avoids these issues: synthetic negatives are generated from in-batch embeddings alone, requiring no external models and no additional data. More on related work in Appendix Sec.~9.

\section{Preliminaries}
\label{sec:preliminaries}

In this section, we present the building blocks for integrating synthetic hard negatives into vision-language contrastive learning. We first review the standard CLIP objective (\cref{sec:clip}), then describe how synthetic negatives augment the contrastive loss in the representation space (\cref{sec:synthetic-hard-negatives}), and finally define the generation strategies (\cref{sec:synthetic-negative-strategies}). These preliminaries set up the vocabulary for \cref{sec:adapting}, where we analyze why naively applying these strategies to the vision-language setting fails and present the design choices that lead to \model.

\begin{figure}[t]
    \centering
    \hspace*{25pt}
    \includegraphics[width=0.98\linewidth]{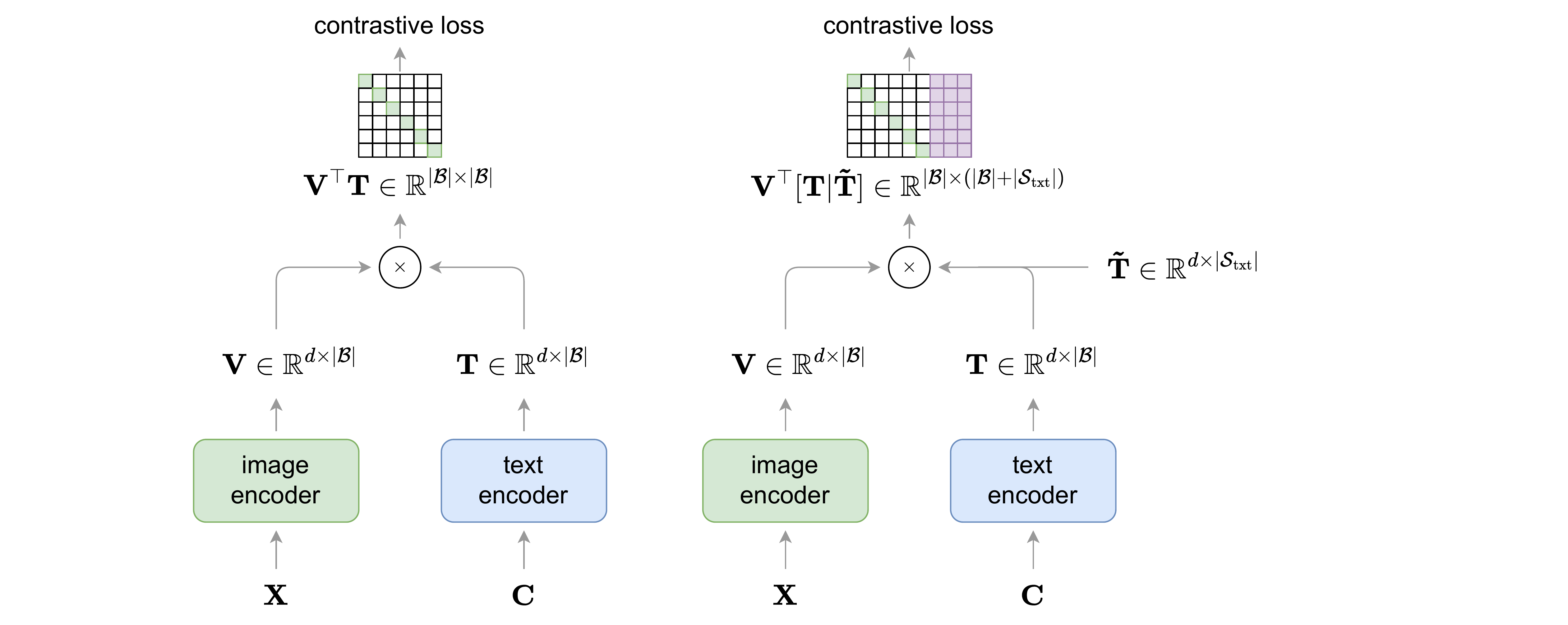}
    \vspace*{-15pt}
    \caption{
    \textbf{Architecture comparison for the image-to-text direction. }
    \textbf{(a)} Standard CLIP encodes images $\mathbf{X}$ and captions $\mathbf{C}$ into embeddings $\mathbf{V} \in \mathbb{R}^{d \times |\mathcal{B}|}$ and $\mathbf{T} \in \mathbb{R}^{d \times |\mathcal{B}|}$, computing the similarity matrix $\mathbf{V}^\top \mathbf{T} \in |\mathcal{B}| \times |\mathcal{B}|$ using only batch negatives. 
    \textbf{(b)} Our method generates synthetic text negatives $\tilde{\mathbf{T}} \in \mathbb{R}^{d \times |\mathcal{S}_{\text{txt}}|}$ directly in the representation space and concatenates them with batch embeddings to form an augmented similarity matrix $\mathbf{V}^\top [\mathbf{T} | \tilde{\mathbf{T}}] \in |\mathcal{B}| \times (|\mathcal{B}| + |\mathcal{S}_{\text{txt}}|)$. 
    In the similarity matrices, {\color{RealPos}green} cells denote positive pairs, white cells denote batch negatives, and {\color{SynNeg}purple} cells denote synthetic negatives. 
    The text-to-image direction follows an analogous procedure with synthetic image negatives (\cf \cref{sec:synthetic-hard-negatives}).
    }
    \label{fig:architecture_comparison}
\end{figure}

\subsection{Contrastive Language-Image Pretraining}
\label{sec:clip}

We assume two parameterized models: an image encoder $f_\theta$ with parameters $\theta$ (\eg, a vision transformer \cite{dosovitskiy2021vit} or a convnet \cite{he2016resnet}) and a text encoder $g_\phi$ with parameters $\phi$ (\eg, a GPT-style transformer \cite{radford2018gpt,radford2019gpt2}). Each input sample $(\mathbf{x}_i, \mathbf{c}_i)$ consists of an image $\mathbf{x}_i$ and text caption $\mathbf{c}_i$. The encoders produce embeddings $\mathbf{v}_i = f_\theta(\mathbf{x}_i)$ and $\mathbf{t}_i = g_\phi(\mathbf{c}_i)$, which are mapped to a common $d$-dimensional embedding space via projection layers and $\ell_2$-normalized, \ie, $\|\mathbf{v}_i\|_2 = \|\mathbf{t}_i\|_2 = 1$. For a batch $\mathcal{B}$, each sample $i$ is associated with one positive pair and $|\mathcal{B}|-1$ in-batch negatives. To aid readability, throughout this section we use {\color{RealNeg} red} to denote batch negatives. The bi-directional contrastive loss is computed as:

\begin{equation}\label{eq:clip-loss}
  \mathcal{L}_{\text{CLIP}} = -\frac{1}{2|\mathcal{B}|}\left(
    \underbrace{\sum_{i \in \mathcal{B}} \log \frac{\exp(\mathbf{v}_i^\top \mathbf{t}_i / \tau)}{{\color{RealNeg}\sum_{j \in \mathcal{B}} \exp(\mathbf{v}_i^\top \mathbf{t}_j / \tau)}}}_{\mathcal{L}_{\text{i2t}}}
    +
    \underbrace{\sum_{i \in \mathcal{B}} \log \frac{\exp(\mathbf{t}_i^\top \mathbf{v}_i / \tau)}{{\color{RealNeg}\sum_{j \in \mathcal{B}} \exp(\mathbf{t}_i^\top \mathbf{v}_j / \tau)}}}_{\mathcal{L}_{\text{t2i}}}
  \right),
\end{equation}

\noindent where $\tau$ is a learnable temperature parameter. We denote the two directional terms as \textit{image-to-text} (i2t) and \textit{text-to-image} (t2i), respectively. The denominators include both the positive term $j = i$ and the $|\mathcal{B}| - 1$ in-batch negatives $j \neq i$, following the standard InfoNCE formulation~\cite{oord2019cpc}. In the standard CLIP formulation~\cite{radford2021clip}, negatives are drawn exclusively from the batch $\mathcal{B}$, requiring large batch sizes (\eg, 32,768) to provide a sufficiently large and diverse pool of negatives. Training with smaller batches significantly limits discriminative ability, as the lack of diverse negatives leads to weaker representations~\cite{chen2020simclr}.

\subsection{Synthetic Hard Negatives in the Representation Space}
\label{sec:synthetic-hard-negatives}

We augment the denominators of both $\mathcal{L}_{\text{i2t}}$ and $\mathcal{L}_{\text{t2i}}$ with synthetically generated hard negatives directly in the representation space. To aid readability, we use {\color{SynNeg} purple} to denote synthetic negatives throughout this section. For the i2t term, we generate a set of synthetic text negatives $\mathcal{S}_{\text{txt}}^i = \{\tilde{\mathbf{t}}_k\}$ for each image query $\mathbf{v}_i$. Similarly, for the t2i term, we generate a set of synthetic image negatives $\mathcal{S}_{\text{img}}^i = \{\tilde{\mathbf{v}}_k\}$ for each text query $\mathbf{t}_i$. The augmented losses are:

\begin{equation}\label{eq:snap-i2t}
  \bar{\mathcal{L}}_{\text{i2t}} = -\frac{1}{|\mathcal{B}|} \sum_{i \in \mathcal{B}} \log \frac{\exp(\mathbf{v}_i^\top \mathbf{t}_i / \tau)}{{\color{RealNeg}\sum_{j \in \mathcal{B}} \exp(\mathbf{v}_i^\top \mathbf{t}_j / \tau)} + {\color{SynNeg}\sum_{\tilde{\mathbf{t}}_k \in \mathcal{S}_{\text{txt}}^i} \exp(\mathbf{v}_i^\top \tilde{\mathbf{t}}_k / \tau)}},
\end{equation}

Each query has its own synthetic-negative set, generated from its
selected hard-negative pool; synthetic negatives are not shared
across queries.

\begin{equation}\label{eq:snap-t2i}
  \bar{\mathcal{L}}_{\text{t2i}} = -\frac{1}{|\mathcal{B}|} \sum_{i \in \mathcal{B}} \log \frac{\exp(\mathbf{t}_i^\top \mathbf{v}_i / \tau)}{{\color{RealNeg}\sum_{j \in \mathcal{B}} \exp(\mathbf{t}_i^\top \mathbf{v}_j / \tau)} + {\color{SynNeg}\sum_{\tilde{\mathbf{v}}_k \in \mathcal{S}_{\text{img}}^i} \exp(\mathbf{t}_i^\top \tilde{\mathbf{v}}_k / \tau)}}.
\end{equation}

\noindent The final loss combines both augmented directional terms:

\begin{equation}\label{eq:snap-loss}
  \bar{\mathcal{L}}_{\text{}} = \frac{1}{2} \left( \bar{\mathcal{L}}_{\text{i2t}} + \bar{\mathcal{L}}_{\text{t2i}} \right).
\end{equation}

\noindent This formulation enriches the negative set with diverse, hard examples that would be computationally prohibitive to generate in the input space, while requiring minimal overhead as the generation occurs directly in the representation space during the forward pass (\cf \cref{sec:computational-efficiency}). With synthetic negatives, we produce high-quality, diverse negatives \textbf{fast} and \textbf{cheaply}, enabling effective training even with moderate batch sizes (\cf Appendix Sec. 9).

\subsection{Synthetic Negative Generation Strategies}
\label{sec:synthetic-negative-strategies}
We define hardness by cosine similarity to the query~\cite{kalantidis2020mochi}. Given a query image embedding $\mathbf{v}_i$, we select the top-$N$ batch text negatives $\hat{\mathcal{T}}_N^i \subset \mathcal{B} \setminus {i}$ with the highest cosine similarity to $\mathbf{v}_i$; analogously, we select $\hat{\mathcal{V}}_N^i$ for text query $\mathbf{t}_i$. We implement six strategies to generate synthetic negatives. For the i2t direction, each strategy produces a synthetic text negative $\tilde{\mathbf{t}}_k$ from the query $\mathbf{v}_i$ and one or two hard negatives $\mathbf{t}_j, \mathbf{t}_l$ sampled uniformly from $\hat{\mathcal{T}}_N^i$

\begin{equation}\label{eq:synthetic_strategies}
{\color{SynNeg}\tilde{\mathbf{t}}_k^{(s)}} = \left\{
\begin{array}{l@{\quad}l@{\quad}l}
\alpha_k \cdot \mathbf{v}_i + (1 - \alpha_k) \cdot {\color{RealNeg}\mathbf{t}_j}, & s = 1, & \alpha_k \sim \mathcal{U}(0, 0.5) \\[4pt]
{\color{RealNeg}\mathbf{t}_j} + \beta_k \cdot ({\color{RealNeg}\mathbf{t}_j} - \mathbf{v}_i), & s = 2, & \beta_k \sim \mathcal{U}(1, 1.5) \\[4pt]
\gamma_k \cdot {\color{RealNeg}\mathbf{t}_j} + (1 - \gamma_k) \cdot {\color{RealNeg}\mathbf{t}_l}, & s = 3, & \gamma_k \sim \mathcal{U}(0, 1) \\[4pt]
{\color{RealNeg}\mathbf{t}_j} + \mathcal{N}(\mathbf{0}, \sigma^2 \cdot \mathbf{I}), & s = 4 & \\[4pt]
{\color{RealNeg}\mathbf{t}_j} + \delta \cdot \nabla_{{\color{RealNeg}\mathbf{t}_j}} (\mathbf{v}_i^\top {\color{RealNeg}\mathbf{t}_j}), & s = 5 & \\[4pt]
{\color{RealNeg}\mathbf{t}_j} + \eta \cdot \text{sign}(\nabla_{{\color{RealNeg}\mathbf{t}_j}} (\mathbf{v}_i^\top {\color{RealNeg}\mathbf{t}_j})), & s = 6 &
\end{array}
\right.
\end{equation}

\noindent All synthetic negatives are $\ell_2$-normalized after generation. Strategies $s{=}1$ (interpolation) and $s{=}2$ (extrapolation) blend the query with hard negatives~\cite{zhang2018mixup}; $s{=}3$ (mixup) combines pairs of hard negatives~\cite{zhang2018mixup,verma2019manifold}; $s{=}4$ perturbs a hard negative with Gaussian noise; and $s{=}5,6$ apply gradient-ascent and sign-based adversarial perturbations~\cite{goodfellow2015fgsm,madry2018pgd} (see~\cite{giakoumoglou2024synco} for further detail). Crucially, $s{=}1$ and $s{=}2$ involve the query embedding $\mathbf{v}_i$ (or $\mathbf{t}_i$ in the t2i direction), producing \textit{cross-modal} synthetic negatives, while $s{=}3$ and $s{=}4$ operate exclusively on same-modality negatives without involving any positive. As we show in \cref{sec:failure-modes}, this distinction is critical in the vision-language setting.  

\section{Adapting Synthetic Negatives to Vision-Language Pretraining}
\label{sec:adapting}

The success of representation-space synthetic negatives in unimodal learning (\cref{sec:synthetic-negative-strategies}) naturally suggests applying the same strategies to CLIP-style training. However, the dual-modality structure of vision-language models introduces failure modes with no unimodal analogue. This raises a central question: \textit{What makes synthetic hard negatives work in vision-language pretraining?} In this section, we systematically analyze these failure modes (\cref{sec:failure-modes}) along their multimodal geometry (\cref{sec:geometry}), their secondary effect on the learnable temperature (\cref{sec:temperature-divergence}), and present the resulting design choices that define \model (\cref{sec:snap-method}). All ablations use CC3M with ViT-B/16, reporting IN-val and IN-v2 zero-shot accuracy. We generate 64 synthetic negatives per query from the top-256 hardest negatives, split equally among active strategies. The hyperparameters $\sigma = 0.01$, $\delta = 0.01$, and $\eta = 0.01$ are set empirically without extensive tuning, following~\cite{giakoumoglou2024synco}.
\subsection{Naive Extension and Failure Modes}
\label{sec:failure-modes}

\begin{figure}[t]
  \centering
  \begin{subfigure}[b]{0.31\linewidth}
    \centering
    \includegraphics[width=\linewidth]{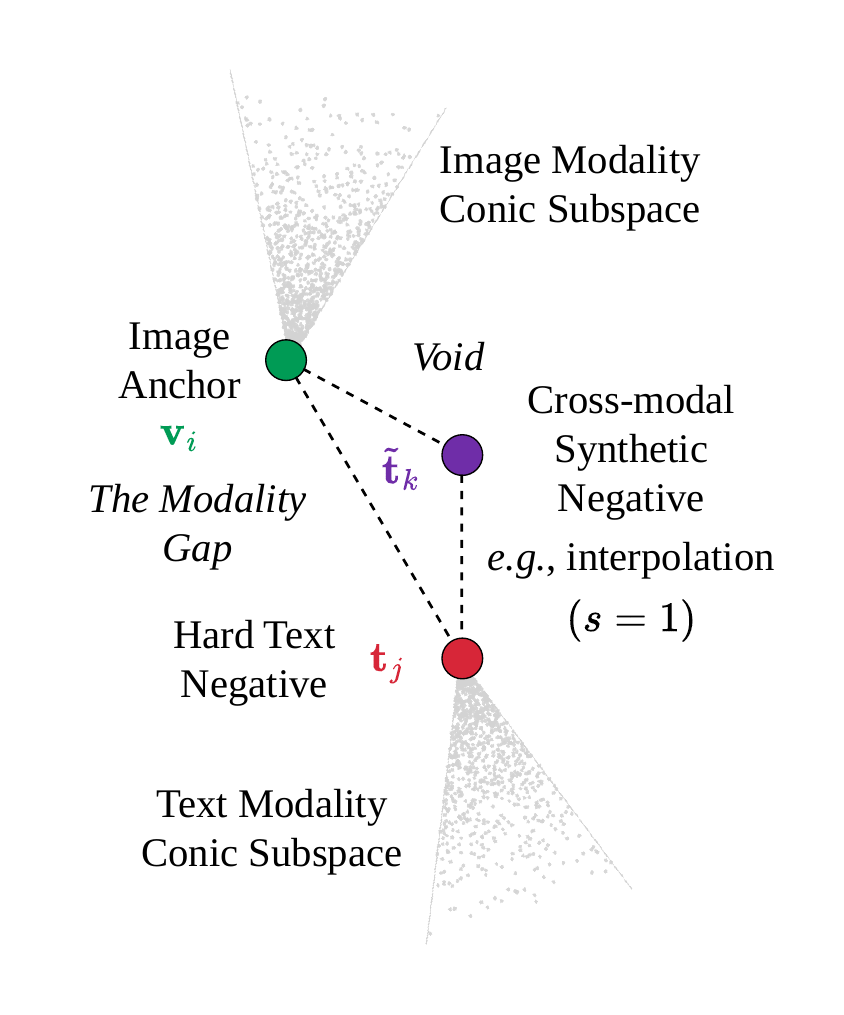}
    \caption{Cross-modal Failure}
    \label{fig:gap}
  \end{subfigure}
  \begin{subfigure}[b]{0.31\linewidth}
    \centering
    \includegraphics[width=\linewidth]{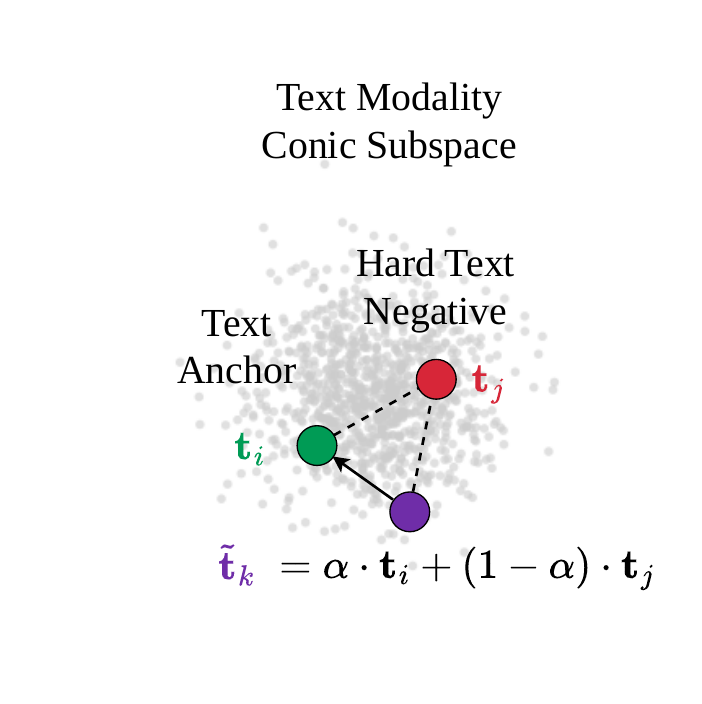}
    \caption{Positive Leakage}
    \label{fig:leakage}
  \end{subfigure}
  \begin{subfigure}[b]{0.31\linewidth}
    \centering
    \includegraphics[width=\linewidth]{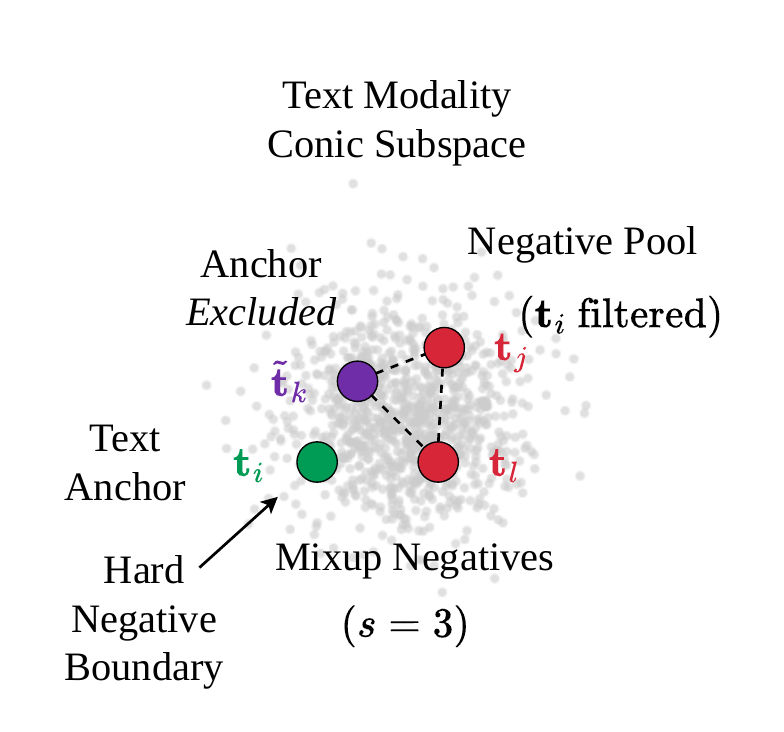}
    \caption{\model (Ours)}
    \label{fig:snap_sol}
  \end{subfigure}
  \caption{\textbf{Geometric intuition of failure modes in vision-language pretraining.} 
  \textbf{(a) Cross-modal failure:} Blending image and text embeddings creates samples in the ``modality gap'' that are trivially rejected (\cref{sec:failure-modes}). 
  \textbf{(b) Positive Leakage:} Including the positive pair in synthesis creates negatives that contain the actual signal, causing gradient conflict (\cref{sec:failure-modes}). 
  \textbf{(c) \model:} By generating intra-modal negatives and filtering the positive anchor, we create high-quality hard negatives that respect the modality geometry (\cref{sec:snap-method}).}
  \label{fig:failure_modes_split}
\end{figure}

\begin{figure}[t]
    \centering
    \begin{minipage}[t]{0.54\textwidth}
        \vspace{0pt}
        \centering
        \captionof{table}{\textbf{Cross-modal vs.\ intra-modal strats.} Applying all six strategies in cross-modal mode degrades performance below the CLIP baseline. Converting to intra-modal mode helps but still underperforms. Restricting to $s=3$ and $s=4$ recovers performance. We {\fboxsep 1pt\colorbox{gray!10}{highlight}} the default hyperparameter.}
        \label{tab:cross-intra}
        \vspace{4pt}
        \begin{tabular}{lcccc}
            \toprule
            \bf Method & \bf Strategies & \bf Mode & \bf IN-val & \bf IN-v2 \\
            \midrule
            CLIP~\cite{radford2021clip} & n/a & n/a & 10.4 & 8.4 \\
            \model & all 6 & cross & 7.6 {\color{red}($\downarrow$)} & 6.8 {\color{red}($\downarrow$)} \\
            \model & all 6 & intra & 8.7 {\color{red}($\downarrow$)} & 7.5 {\color{red}($\downarrow$)} \\
            \model & \cellcolor{gray!10}$s{=}3,4$ & \cellcolor{gray!10}intra & 10.0 {\color{red}($\downarrow$)} & 8.0 {\color{red}($\downarrow$)} \\
            \bottomrule
        \end{tabular}
    \end{minipage}
    \hfill
    \begin{minipage}[t]{0.44\textwidth}
        \vspace{0pt}
        \centering
        \includegraphics[width=0.9\linewidth]{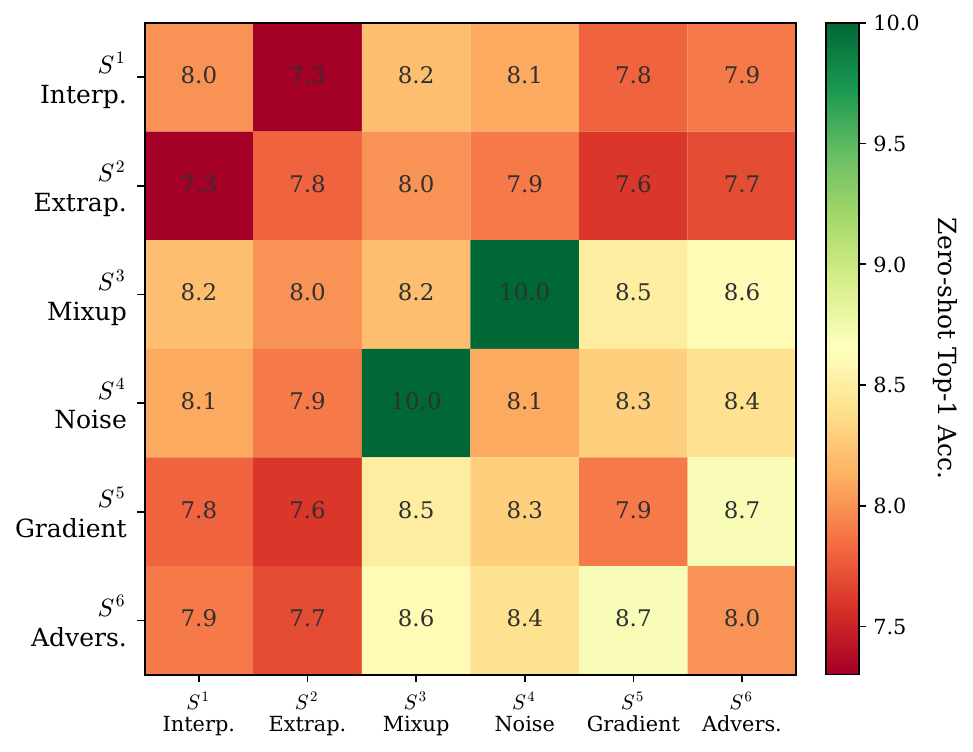}
        \captionof{figure}{\textbf{Pairwise strategy complementarity on IN-val}. Strategies $s=3,4$ achieve the highest accuracy.}
        \label{fig:pairwise_inval}
    \end{minipage}
\end{figure}

\subsubsection{Cross-modal synthetic negatives fail due to geometry.}

The most direct extension is to apply all six strategies from \cref{eq:synthetic_strategies} as defined, where $s{=}1$ and $s{=}2$ blend the image query $\mathbf{v}_i$ with text negatives $\mathbf{t}_j$. However, vision-language models exhibit a well-documented \textit{modality gap}~\cite{liang2022mind} (\cref{fig:gap}): image and text embeddings cluster in distinct cones of the shared space, and cross-modal interpolations such as $\alpha \cdot \mathbf{v}_i + (1 - \alpha) \cdot \mathbf{t}_j$ produce vectors in the void between them that real embeddings never occupy, making them trivially separable. As shown in \Cref{tab:cross-intra}, restricting to $s = 3, 4$ in intra-modal mode nearly recovers the CLIP baseline (10.0 vs.\ 10.4 on IN-val); as we show in \Cref{sec:temperature-divergence}, the remaining gap is closed and surpassed once the temperature is fixed.

\subsubsection{Intra-modal negatives involving the positive fail due to semantic leakage.}

A natural fix is to convert all strategies to intra-modal variants by replacing the cross-modal query $\mathbf{v}_i$ in $s = 1, 2$ with the matched text positive $\mathbf{t}_i$. However, this introduces a subtler failure. Consider the intra-modal interpolation: $\tilde{\mathbf{t}} = \alpha \cdot \mathbf{t}_i + (1 - \alpha) \cdot \mathbf{t}_j$, where $\mathbf{t}_i$ is the positive caption for query $\mathbf{v}_i$ and $\mathbf{t}_j$ is a different caption. Because $\mathbf{t}_i$ is the correct match for $\mathbf{v}_i$, the synthetic ``negative'' $\tilde{\mathbf{t}}$ inherits positive signal. As $\alpha \to 1$, it becomes nearly identical to the true positive, yet the InfoNCE denominator treats it as a negative. This \textit{positive leakage} (\cref{fig:leakage}) sends contradictory gradients: the numerator of \cref{eq:snap-i2t} encourages similarity with $\mathbf{t}_i$, while the denominator penalizes similarity with $\tilde{\mathbf{t}} \approx \mathbf{t}_i$. \cref{tab:cross-intra} confirms that intra-modal generation with all six strategies (8.7 on IN-val) improves over the cross-modal variant but still underperforms the CLIP baseline.

\subsubsection{Two distinct failure modes, one solution.}
The tested constructions alter negative geometry through cross-modal
operations or introduce positive signal through explicit inclusion
of the matched positive. Strategies $s \in \{3, 4\}$ avoid both by generating negatives from non-matching pairs only. Restricting to $s \in \{3, 4\}$ nearly recovers the CLIP baseline (10.0 vs.\ 10.4 on IN-val, 8.0 vs.\ 8.4 on IN-v2; \Cref{tab:cross-intra}), with the remaining gap closed  and surpassed once temperature is fixed (\Cref{sec:temperature-divergence}). \Cref{fig:pairwise_inval} confirms this: $s=3$ and $s=4$ consistently outperform any combination involving cross-modal blending or the positive pair.

\subsubsection{Temperature divergence.}
\label{sec:temperature-divergence}

\begin{figure}[t]
    \centering
    \begin{minipage}[t]{\textwidth}
        \centering
        \begin{minipage}[t]{0.48\textwidth}
            \vspace{0pt}
            \centering
            \captionof{table}{\textbf{Temperature ablation.} F\textbf{Temperature ablation.} With fixed $\tau=0.07$, CLIP+\model outperforms both fixed- and learnable-temperature CLIP. We {\fboxsep 1pt\colorbox{gray!10}{highlight}} the default setting.}
            \label{tab:temperature}
            \vspace{4pt}
            \begin{tabular}{lccc}
                \toprule
                \bf Method & $\boldsymbol{\tau}$ & \bf IN-val & \bf IN-v2 \\
                \midrule
                CLIP & learn. & 10.4 & 8.4 \\
                CLIP & fixed  & 10.1 & 8.0 \\
                CLIP+\model & learn. & $10.0\,{\color{red}(\downarrow)}$ & $8.0\,{\color{red}(\downarrow)}$ \\
                CLIP+\model & \cellcolor{gray!10}fixed & $\mathbf{11.0}\,{\color{green}(\uparrow)}$ & $\mathbf{9.6}\,{\color{green}(\uparrow)}$ \\
                \bottomrule
            \end{tabular}
        \end{minipage}
        \hfill
        \begin{minipage}[t]{0.48\textwidth}
            \vspace{0pt}
            \centering
            \includegraphics[width=0.85\linewidth]{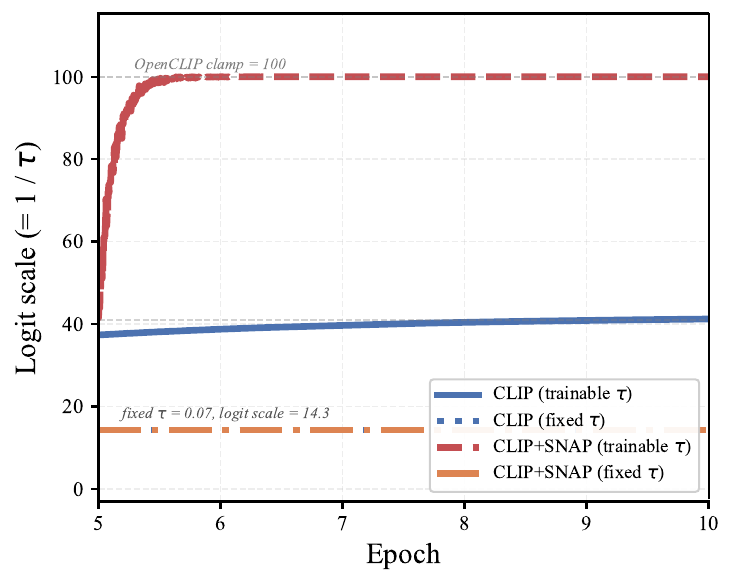}
            \captionof{figure}{\textbf{Logit scale trajectories} (ViT-B/32, CC3M, epochs 5 to 10). CLIP+SNAP with a learnable $\tau$ saturates the OpenCLIP clamp within half an epoch. Both fixed-$\tau$ runs remain stable.}
            \label{fig:temperature}
        \end{minipage}
    \end{minipage}

    \vspace{10pt}

    \begin{minipage}[t]{\textwidth}
        \centering
        \begin{minipage}[t]{0.48\textwidth}
            \vspace{0pt}
            \centering
            \includegraphics[width=\linewidth]{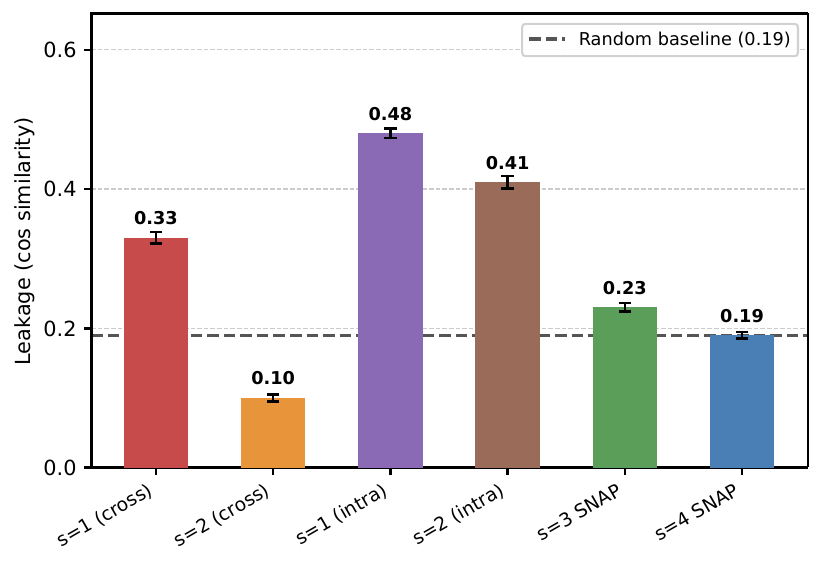}
            \captionof{figure}{\textbf{Positive leakage per strategy} (ViT-B/32, CC3M). mean $\cos(\tilde{\mathbf{t}}, \mathbf{t}_i)$ per strategy; dashed line marks the random baseline ($0.19$).}
            \label{fig:pos_leakage}
        \end{minipage}
        \hfill
        \begin{minipage}[t]{0.48\textwidth}
            \vspace{0pt}
            \centering
            \captionof{table}{\textbf{TSR and relative hardness.} Lower TSR and higher $\Delta$ hardness indicate harder, more informative negatives. We \colorbox{gray!10}{highlight} SNAP strategies.}
            \label{tab:tsr}
            \vspace{4pt}
            \begin{tabular}{lcc}
                \toprule
                \bf Strategy & \bf TSR\,(\%)$\downarrow$ & \bf $\Delta$ hardness$\uparrow$ \\
                \midrule
                $s{=}1$ (cross)  &  0.3 & $+0.281$ \\
                $s{=}2$ (cross)  & 87.8 & $-0.189$ \\
                $s{=}1$ (intra)  &  0.6 & $+0.133$ \\
                $s{=}2$ (intra)  & 74.0 & $-0.096$ \\
                \midrule
                \cellcolor{gray!10}$s{=}3$ (SNAP) & 2.2 & $+0.041$ \\
                \cellcolor{gray!10}$s{=}4$ (SNAP) & 13.2 & $+0.045$ \\
                \bottomrule
            \end{tabular}
        \end{minipage}
    \end{minipage}
\end{figure}

Adding synthetic hard negatives changes the contrastive denominator and its interaction with the learnable temperature $\tau$. In the evaluated configuration, CLIP+\model with learnable $\tau$ reaches the OpenCLIP logit-scale clamp of 100 within half an epoch after synthetic negatives are enabled (\cref{fig:temperature}). This occurs with the positive-free intra-modal strategies $s=3,4$. Fixing $\tau=0.07$ improves IN-val accuracy from 10.0 to 11.0 and IN-v2 accuracy from 8.0 to 9.6 (\cref{tab:temperature}). Under the same fixed temperature, CLIP without synthetic negatives achieves 10.1 and 8.0, respectively. We therefore adopt fixed $\tau=0.07$ as part of the \model training recipe. The two cross-modal strategies fail through distinct mechanisms: extrapolation ($s{=}2$) pushes vectors beyond the text cone, while interpolation ($s{=}1$) pulls toward the image query and thus the positive, making positive leakage the dominant failure rather than trivial separability, as we quantify in \Cref{sec:geometry}.

\subsection{Multimodal Geometry of Synthetic Negatives}
\label{sec:geometry}

We characterize the failure modes in \cref{sec:failure-modes} using four complementary representation level diagnostics computed on ViT-B/32 trained on CC3M.

\subsubsection{Leakage, separability, and hardness diagnostics.}
We quantify the failure modes with three metrics over 2,000 validation pairs (ViT-B/32, CC3M): \textit{positive leakage} $\cos(\tilde{\mathbf{t}}, \mathbf{t}_i)$, where values above the random baseline ($0.19$) indicate semantic contamination; \textit{trivially separable rate} (TSR), the fraction of synthetics easier than the 10th percentile of real hard negatives; and \textit{relative hardness} $\Delta = \cos(\mathbf{v}_i, \tilde{\mathbf{t}}) - \cos(\mathbf{v}_i, \mathbf{t}_\text{rand})$. As shown in \cref{fig:pos_leakage,tab:tsr}, cross-modal strategies fail for opposite reasons: $s{=}2$ (cross) lands in the void beyond the text cone (TSR $87.8\%$, $\Delta = -0.189$), while $s{=}1$ (cross) interpolates toward the positive (leakage $0.33$). Intra-modal strategies involving the positive suffer from contamination ($s{=}1$ intra: $0.48$; $s{=}2$ intra: $0.41$) or trivial separability ($s{=}2$ intra: TSR $74.0\%$). Only $s{=}3$ and $s{=}4$ avoid cross-modal mixing and explicit positive inclusion, exhibiting leakage close to the random baseline ($0.23$ and $0.19$), low TSR ($2.2\%$ and $13.2\%$), and positive relative hardness. 

\subsubsection{PCA subspace visualization.}
\begin{figure}[t]
    \centering
    \includegraphics[width=\linewidth]{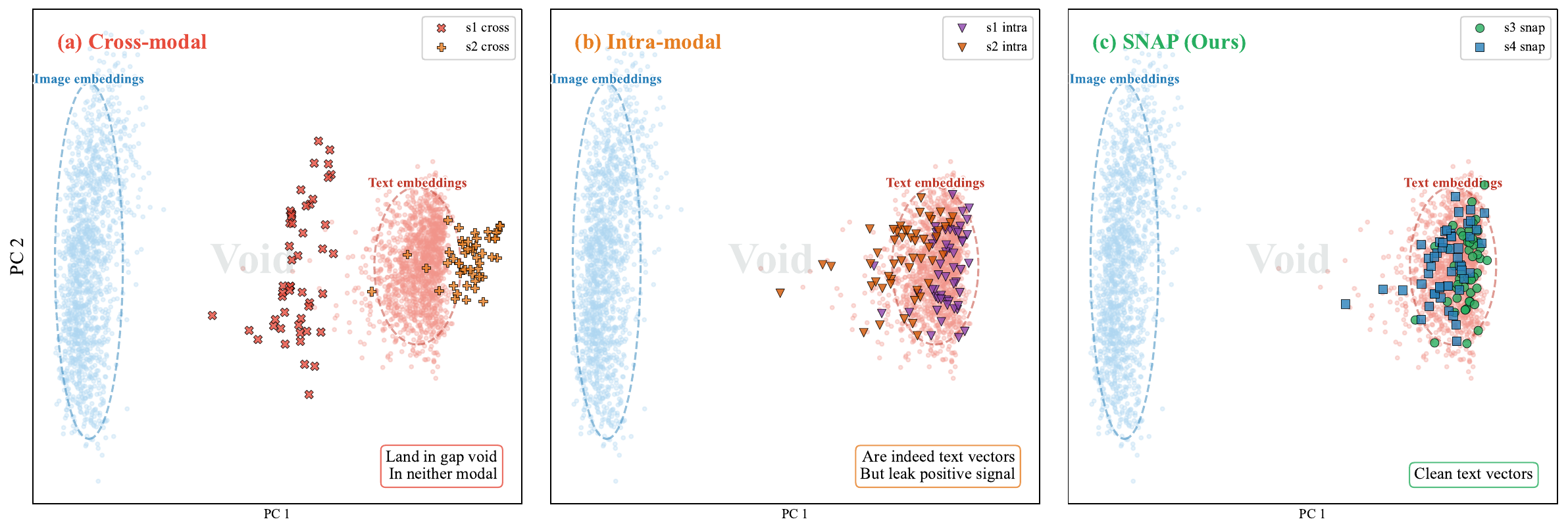}
    \caption{\textbf{PCA subspace  of synthetic negatives} (ViT-B/32, CC3M). \textbf{(a)} Cross-modal negatives land in the gap void between modality cones. \textbf{(b)} Intra-modal negatives involving the positive cluster near the positive anchor, causing leakage. \textbf{(c)} SNAP negatives ($s{=}3$, $s{=}4$) are confined within the text cone and away from the positive.}
    \label{fig:pca}
\end{figure}

\cref{fig:pca} projects image embeddings, text embeddings, and synthetic negatives onto the first two principal components. Cross-modal strategies scatter into the void between the image and text cones, a region real embeddings never occupy. Intra-modal strategies involving the positive cluster within the text cone but near the positive anchor, explaining the observed leakage. SNAP strategies ($s{=}3$, $s{=}4$) are confined within the text cone and distributed away from the positive, occupying the same subspace as real text negatives.

\subsection{\model}
\label{sec:snap-method}

Guided by the analysis in \cref{sec:failure-modes,sec:geometry}, we propose \model (\underline{\textbf{S}}ynthetically \underline{\textbf{N}}egative \underline{\textbf{A}}ugmented \underline{\textbf{P}}retraining) that makes three principled design choices.

\begin{wraptable}{r}{0.48\textwidth}
    \vspace{-42pt}
    \centering
    \caption{\textbf{Synthesis and directionality ablations on CC3M.}}
\label{tab:synthesis-direction}
\begin{tabular}{@{}lcc@{}}
    \toprule
    Variant & IN-val & IN-v2 \\
    \midrule
    Random-neg.\ dup.       & 10.0 & 8.0 \\
    Hard-neg.\ dup.         & 10.4 & 8.5 \\
    Hard-neg.\ wt.    & 10.5 & 8.7 \\
    $s=3$          & 10.8 & 9.2 \\
    $s=4$          & 10.6 & 8.8 \\
    \midrule
    $s=3,4$ (i2t only)       & 10.7 & 8.8 \\
    $s=3,4$ (t2i only)       & 10.6 & 8.7 \\
    \midrule
    \model        & \textbf{11.0} & \textbf{9.6} \\
    \bottomrule
\end{tabular}
\vspace{-42pt}
\end{wraptable}

\subsubsection{Intra-modal, positive-free strategies.} \model uses only strategies $s = 3$ and $s = 4$  from \cref{eq:synthetic_strategies}, both operating in intra-modal mode (\cref{fig:snap_sol}). These avoid cross-modal geometric collapse and intra-modal positive leakage (\cref{tab:cross-intra,fig:pairwise_inval}).

\subsubsection{Fixed temperature.} \model fixes the temperature $\tau$ rather than learning it, preventing the logit scale saturation that arises when hard negatives interact with a learnable temperature (\cref{tab:temperature}).

\subsubsection{Bidirectional synthesis.}
\model generates synthetic negatives for both i2t and t2i directions. We compare against repeating randomly sampled or hard negatives in the contrastive denominator, and increasing the weight of hard-negative terms. Both synthesis strategies outperform these controls, with their combination performing best (\cref{tab:synthesis-direction}). Bidirectional synthesis also outperforms either direction alone.

\subsubsection{Computational overhead.}
\label{sec:computational-efficiency}

Since \model generates negatives via vector operations—specifically additions, scalar multiplications, and normalization—on pre-computed embeddings, it avoids additional forward or backward passes through the encoders. This design ensures minimal impact on training efficiency, requiring no external generative models or architectural modifications. On a ViT-B/16 backbone, CLIP \cite{radford2021clip} requires 32.5 minutes per epoch, while \model adds only 3 minutes, resulting in an 9.23\% overhead. Similarly, FLIP \cite{li2023flip} requires 14.9 minutes per epoch, and the addition of \model increases this to 16.2 minutes, a comparable overhead of 8.72\%. These results demonstrate that \model provides significant discriminative gains with a negligible increase in training time. Further ablations on batch size and the number of synthetic negatives are provided in the supplementary material (Appendix Sec.~10).

\section{Experiments}
\label{sec:experiments}

We train the CLIP and FLIP baselines and their \model variants under matched data and training configurations, except for the temperature treatment described above. We also train SigLIP~\cite{zhai2023siglip} under our training setup. Results for LaCLIP~\cite{fan2023laclip}, TripletCLIP~\cite{patel2024tripletclip}, and NegCLIP~\cite{yuksekgonul2023negclip} are taken from~\cite{patel2024tripletclip} and included for reference only. Detailed implementation details in Appendix Sec.~8.
\subsection{Implementation Details}
\label{sec:implementation-details}

\subsubsection{Pretraining datasets.} We pretrain on two image-text datasets of varying scale: CC3M~\cite{cc3m} and CC12M~\cite{cc12m}. Due to image link rot, the versions we obtained contain fewer samples than the original releases: $\approx$1.7M out of 3.3M for CC3M and $\approx$7.2M out of 12.4M for CC12M. This may lead to minor performance differences compared to models trained on the full datasets.

\subsubsection{Model architectures.} For the image encoder $f_\theta$, we use ViT-B/16, ViT-B/32~\cite{dosovitskiy2021vit}, and ResNet-50~\cite{he2016resnet}. For the text encoder $g_\phi$, we use a 12-layer Transformer~\cite{radford2021clip}. Both encoders include projection heads mapping to a shared $d = 512$ dimensional embedding space.

\subsubsection{Training setup.} We use AdamW~\cite{adamw} with learning rate $5 \times 10^{-4}$, weight decay $0.2$, and cosine scheduling with 10k warmup steps. Training uses batch size $|\mathcal{B}| = 4096$ on 4$\times$ A100 GPUs for 32 epochs, with $224 \times 224$ images and text length 77. The temperature $\tau$ is initialized to $0.07$ (and kept constant for \model).

\subsubsection{Reproducibility.} For CC3M, we pretrain three checkpoints (different seeds) and report the mean, with standard deviations as subscripts in \Cref{tab:retrieval,tab:zeroshot,tab:linear}; for CC12M, a single checkpoint is used due to computational constraints.


\clearpage
\subsection{Zero-Shot Image-Text Retrieval}
\label{sec:results-retrieval}

We evaluate cross-modal retrieval on Flickr30k~\cite{flickr30k} and MSCOCO~\cite{mscoco}, computing pairwise cosine similarities over the full test split and reporting Recall@\{1,5,10\} for both image-to-text and text-to-image retrieval in \Cref{tab:retrieval}. \model improves retrieval in the large majority of settings for both CLIP and FLIP, holding across all three backbones and both pretraining scales, with gains appearing on both retrieval directions rather than favoring one; the few exceptions are isolated cells that are unchanged or lower by at most a few tenths of a point.

\begin{table*}[!tbhp]
\centering
\scriptsize
\caption{\textbf{Zero-shot image-text retrieval on Flickr30k and MSCOCO.} We report Recall@K (R@K) metrics for text and image retrieval. \textbf{Symbols}: $^{\ddagger}$ Taken from~\cite{patel2024tripletclip}}
\label{tab:retrieval}
\setlength{\tabcolsep}{0.5pt}
\begin{tabular}{llcccccccccccc}
\toprule
\multirow{3}{*}{\textbf{Data}} & \multirow{3}{*}{\textbf{Method}} & \multicolumn{6}{c}{\textbf{Text Retrieval}} & \multicolumn{6}{c}{\textbf{Image Retrieval}} \\
\cmidrule(lr){3-8} \cmidrule(lr){9-14}
& & \multicolumn{3}{c}{Flickr30k} & \multicolumn{3}{c}{MSCOCO} & \multicolumn{3}{c}{Flickr30k} & \multicolumn{3}{c}{MSCOCO} \\
\cmidrule(lr){3-5} \cmidrule(lr){6-8} \cmidrule(lr){9-11} \cmidrule(lr){12-14}
& & R@1 & R@5 & R@10 & R@1 & R@5 & R@10 & R@1 & R@5 & R@10 & R@1 & R@5 & R@10 \\
\midrule
\multicolumn{14}{c}{{ViT-B/32}} \\
\multirow{8}{*}{\rotatebox{90}{CC3M}}
& SigLIP \cite{zhai2023siglip} & 7.2\std{2} & 19.8\std{3} & 26.5\std{1} & 3.1\std{3} & 10.0\std{1} & 15.7\std{2} & 5.3\std{1} & 15.2\std{2} & 22.3\std{1} & 2.8\std{1} & 9.0\std{1} & 14.0\std{1} \\
& NegCLIP \cite{yuksekgonul2023negclip}$^{\ddagger}$ & 4.9 & 13.8 & 19.6 & 1.7 & 6.3 & 10.4 & 4.8 & 13.0 & 18.5 & 2.0 & 6.6 & 10.6 \\
& LaCLIP \cite{fan2023laclip}$^{\ddagger}$ & 3.7 & 10.9 & 16.0 & 1.6 & 5.1 & 8.9 & 3.5 & 10.8 & 16.0 & 1.7 & 6.0 & 9.5 \\
& TripletCLIP \cite{patel2024tripletclip}$^{\ddagger}$ & 9.1 & 22.0 & 29.8 & 3.2 & 10.4 & 16.2 & 8.4 & 22.0 & 29.6 & 3.6 & 11.3 & 17.4 \\
& CLIP \cite{radford2021clip} & 9.8\std{2} & 25.5\std{3} & 35.7\std{2} & 5.0\std{2} & 15.0\std{1} & 22.3\std{1} & 6.9\std{1} & 19.8\std{2} & 28.5\std{1} & 4.2\std{2} & 12.7\std{2} & 19.0\std{1} \\
& \quad + \model & \cellcolor{blue!10}10.4\std{2} & \cellcolor{blue!10}26.3\std{1} & \cellcolor{blue!10}36.9\std{3} & \cellcolor{blue!10}5.5\std{2} & \cellcolor{blue!10}15.2\std{2} & \cellcolor{blue!10}23.0\std{2} & \cellcolor{blue!10}7.9\std{1} & \cellcolor{blue!10}20.4\std{2} & \cellcolor{blue!10}28.9\std{1} & \cellcolor{blue!10}4.1\std{1} & \cellcolor{blue!10}13.6\std{1} & \cellcolor{blue!10}20.1\std{2} \\
& FLIP \cite{li2023flip} & 6.3\std{3} & 18.5\std{1} & 26.4\std{2} & 3.3\std{3} & 10.7\std{3} & 16.7\std{3} & 5.8\std{3} & 16.9\std{1} & 24.0\std{3} & 3.1\std{1} & 9.6\std{1} & 14.7\std{2} \\
& \quad + \model & \cellcolor{blue!10}6.8\std{1} & \cellcolor{blue!10}19.0\std{1} & \cellcolor{blue!10}26.9\std{3} & \cellcolor{blue!10}3.7\std{1} & \cellcolor{blue!10}11.1\std{3} & \cellcolor{blue!10}17.1\std{1} & \cellcolor{blue!10}6.3\std{1} & \cellcolor{blue!10}17.4\std{3} & \cellcolor{blue!10}24.5\std{1} & \cellcolor{blue!10}3.5\std{1} & \cellcolor{blue!10}10.0\std{2} & \cellcolor{blue!10}15.1\std{2} \\
\midrule
\multirow{8}{*}{\rotatebox{90}{CC12M}}
& SigLIP \cite{zhai2023siglip} & 31.4 & 57.2 & 69.5 & 16.2 & 36.1 & 47.2 & 23.1 & 48.8 & 60.8 & 11.5 & 28.1 & 38.4 \\
& NegCLIP \cite{yuksekgonul2023negclip}$^{\ddagger}$ & 24.7 & 46.6 & 58.2 & 12.3 & 30.2 & 41.4 & 18.7 & 41.7 & 53.4 & 8.6 & 23.1 & 32.6 \\
& LaCLIP \cite{fan2023laclip}$^{\ddagger}$ & 21.3 & 42.7 & 54.6 & 10.5 & 25.9 & 35.6 & 15.1 & 36.3 & 47.0 & 7.2 & 19.8 & 28.4 \\
& TripletCLIP \cite{patel2024tripletclip}$^{\ddagger}$ & 28.0 & 55.9 & 65.7 & 14.6 & 33.0 & 43.8 & 25.3 & 52.4 & 63.3 & 11.4 & 28.5 & 39.0 \\
& CLIP \cite{radford2021clip} & 33.1 & 59.5 & 71.2 & 17.8 & 39.4 & 50.3 & 24.2 & 50.9 & 63.0 & 13.1 & 30.5 & 41.2 \\
& \quad + \model & \cellcolor{blue!10}33.8 & \cellcolor{blue!10}60.0 & \cellcolor{blue!10}71.9 & \cellcolor{blue!10}18.6 & \cellcolor{blue!10}39.8 & \cellcolor{blue!10}51.3 & \cellcolor{blue!10}24.9 & \cellcolor{blue!10}51.4 & \cellcolor{blue!10}63.5 & \cellcolor{blue!10}13.7 & \cellcolor{blue!10}31.0 & \cellcolor{blue!10}42.6 \\
& FLIP \cite{li2023flip} & 26.2 & 50.1 & 61.8 & 14.4 & 32.1 & 43.5 & 20.2 & 44.1 & 56.1 & 10.7 & 26.2 & 36.4 \\
& \quad + \model & \cellcolor{blue!10}27.2 & \cellcolor{blue!10}50.9 & \cellcolor{blue!10}63.1 & \cellcolor{blue!10}15.1 & \cellcolor{blue!10}32.7 & \cellcolor{blue!10}44.5 & \cellcolor{blue!10}20.9 & \cellcolor{blue!10}45.4 & \cellcolor{blue!10}56.8 & \cellcolor{blue!10}10.4 & \cellcolor{blue!10}26.8 & \cellcolor{blue!10}37.2 \\
\midrule
\multicolumn{14}{c}{{ViT-B/16}} \\
\multirow{5}{*}{\rotatebox{90}{CC3M}}
& SigLIP \cite{zhai2023siglip} & 13.6\std{3} & 31.5\std{1} & 43.6\std{1} & 7.0\std{2} & 19.8\std{1} & 28.1\std{2} & 9.3\std{1} & 24.5\std{2} & 33.4\std{1} & 5.6\std{1} & 15.8\std{2} & 23.1\std{3} \\
& CLIP \cite{radford2021clip} & 16.2\std{3} & 37.8\std{2} & 49.6\std{2} & 7.7\std{2} & 21.5\std{2} & 31.5\std{2} & 12.0\std{2} & 28.4\std{2} & 38.2\std{1} & 6.6\std{3} & 18.1\std{3} & 26.2\std{1} \\
& \quad + \model & \cellcolor{blue!10}17.1\std{3} & \cellcolor{blue!10}38.7\std{2} & \cellcolor{blue!10}50.5\std{1} & \cellcolor{blue!10}8.1\std{1} & \cellcolor{blue!10}22.1\std{3} & \cellcolor{blue!10}32.7\std{3} & \cellcolor{blue!10}12.8\std{1} & \cellcolor{blue!10}28.9\std{1} & \cellcolor{blue!10}39.2\std{2} & \cellcolor{blue!10}7.2\std{2} & \cellcolor{blue!10}18.4\std{1} & \cellcolor{blue!10}26.8\std{1} \\
& FLIP \cite{li2023flip} & 12.1\std{2} & 31.2\std{2} & 42.1\std{1} & 6.4\std{2} & 17.7\std{1} & 26.4\std{1} & 10.0\std{2} & 24.8\std{1} & 33.6\std{3} & 5.1\std{1} & 15.4\std{1} & 22.7\std{2} \\
& \quad + \model & \cellcolor{blue!10}12.3\std{2} & \cellcolor{blue!10}31.7\std{3} & \cellcolor{blue!10}43.1\std{1} & \cellcolor{blue!10}7.1\std{3} & \cellcolor{blue!10}18.9\std{2} & \cellcolor{blue!10}27.0\std{3} & \cellcolor{blue!10}10.9\std{3} & \cellcolor{blue!10}25.6\std{3} & \cellcolor{blue!10}34.8\std{1} & \cellcolor{blue!10}5.9\std{1} & \cellcolor{blue!10}16.3\std{3} & \cellcolor{blue!10}24.0\std{2} \\
\midrule
\multirow{5}{*}{\rotatebox{90}{CC12M}}
& SigLIP \cite{zhai2023siglip} & 49.4 & 74.2 & 83.5 & 26.1 & 50.4 & 62.1 & 34.2 & 61.9 & 73.1 & 19.1 & 40.5 & 52.1 \\
& CLIP \cite{radford2021clip} & 53.6 & 77.5 & 85.9 & 29.4 & 53.8 & 65.8 & 38.5 & 66.2 & 76.9 & 21.8 & 44.4 & 56.1 \\
& \quad + \model & \cellcolor{blue!10}55.1 & \cellcolor{blue!10}78.3 & \cellcolor{blue!10}87.1 & \cellcolor{blue!10}30.2 & \cellcolor{blue!10}54.7 & \cellcolor{blue!10}67.1 & \cellcolor{blue!10}39.0 & \cellcolor{blue!10}67.6 & \cellcolor{blue!10}78.2 & \cellcolor{blue!10}22.3 & \cellcolor{blue!10}45.0 & \cellcolor{blue!10}56.6 \\
& FLIP \cite{li2023flip} & 47.9 & 73.2 & 82.2 & 25.5 & 49.1 & 61.2 & 33.6 & 60.5 & 72.1 & 18.8 & 39.4 & 50.9 \\
& \quad + \model & \cellcolor{blue!10}48.5 & \cellcolor{blue!10}74.1 & \cellcolor{blue!10}82.3 & \cellcolor{blue!10}26.4 & \cellcolor{blue!10}49.6 & \cellcolor{blue!10}62.2 & \cellcolor{blue!10}34.3 & \cellcolor{blue!10}61.0 & \cellcolor{blue!10}72.7 & \cellcolor{blue!10}19.5 & \cellcolor{blue!10}39.9 & \cellcolor{blue!10}52.1 \\
\midrule
\multicolumn{14}{c}{{RN-50}} \\
\multirow{5}{*}{\rotatebox{90}{CC3M}}
& SigLIP \cite{zhai2023siglip} & 19.4\std{1} & 44.8\std{1} & 56.4\std{2} & 10.5\std{1} & 27.0\std{1} & 37.5\std{3} & 14.0\std{2} & 31.5\std{2} & 42.0\std{1} & 8.0\std{1} & 21.3\std{1} & 29.8\std{1} \\
& CLIP \cite{radford2021clip} & 20.8\std{3} & 45.8\std{3} & 56.8\std{1} & 11.7\std{3} & 29.6\std{3} & 41.4\std{1} & 16.1\std{2} & 36.5\std{2} & 47.1\std{1} & 9.6\std{1} & 24.3\std{2} & 34.0\std{2} \\
& \quad + \model & \cellcolor{blue!10}21.1\std{2} & \cellcolor{blue!10}47.2\std{1} & \cellcolor{blue!10}58.1\std{1} & \cellcolor{blue!10}11.6\std{1} & \cellcolor{blue!10}29.4\std{1} & \cellcolor{blue!10}42.1\std{1} & \cellcolor{blue!10}16.0\std{1} & \cellcolor{blue!10}37.4\std{1} & \cellcolor{blue!10}47.8\std{1} & \cellcolor{blue!10}10.9\std{1} & \cellcolor{blue!10}25.3\std{3} & \cellcolor{blue!10}34.9\std{3} \\
& FLIP \cite{li2023flip} & 20.6\std{1} & 44.2\std{2} & 56.5\std{2} & 11.3\std{1} & 29.4\std{2} & 40.4\std{1} & 15.8\std{2} & 36.0\std{1} & 47.0\std{3} & 8.8\std{2} & 23.6\std{1} & 33.1\std{3} \\
& \quad + \model & \cellcolor{blue!10}21.2\std{1} & \cellcolor{blue!10}44.6\std{2} & \cellcolor{blue!10}57.2\std{1} & \cellcolor{blue!10}12.2\std{3} & \cellcolor{blue!10}29.9\std{3} & \cellcolor{blue!10}40.9\std{2} & \cellcolor{blue!10}16.2\std{3} & \cellcolor{blue!10}36.9\std{1} & \cellcolor{blue!10}46.9\std{2} & \cellcolor{blue!10}9.2\std{3} & \cellcolor{blue!10}24.0\std{2} & \cellcolor{blue!10}34.5\std{2} \\
\bottomrule
\end{tabular}
\end{table*}

\clearpage
\subsection{Zero-Shot Classification}
\label{sec:results-zeroshot}

We evaluate zero-shot classification on ImageNet~\cite{imagenet,imagenet-ilsvrc} and 10 downstream datasets spanning fine-grained recognition (Stanford Cars~\cite{stanfordcars}, FGVC Aircraft~\cite{aircraft}), scene recognition (SUN397~\cite{sun}), texture classification (DTD~\cite{dtd}), and generic object classification (CIFAR~\cite{cifar10}), encoding class names as text prompts and predicting via cosine similarity without any task-specific training (\Cref{tab:zeroshot}). \model improves accuracy across settings, with the clearest gains on the fine-grained Cars and Aircraft benchmarks, where CC3M ViT-B/32 accuracy roughly doubles (\eg, Cars $0.8 \to 1.9$, Aircraft $1.9 \to 3.4$ for CLIP), while general benchmarks such as ImageNet show steady improvements.

\begin{table*}[!tbhp]
\centering
\scriptsize
\caption{\textbf{Zero-shot classification.} Top-1 accuracy on ImageNet and 10 downstream datasets spanning fine-grained, scene, and object recognition. $^{\ddagger}$Taken from~\cite{patel2024tripletclip}.}
\label{tab:zeroshot}
\setlength{\tabcolsep}{1.7pt}
\begin{tabular}{llccccccccccc}
\toprule
\textbf{Data} & \textbf{Model} &
\rotatebox[origin=lb]{90}{\smash{\textbf{Food-101}}} &
\rotatebox[origin=lb]{90}{\smash{\textbf{CIFAR-10}}} &
\rotatebox[origin=lb]{90}{\smash{\textbf{CIFAR-100}}} &
\rotatebox[origin=lb]{90}{\smash{\textbf{SUN397}}} &
\rotatebox[origin=lb]{90}{\smash{\textbf{Cars}}} &
\rotatebox[origin=lb]{90}{\smash{\textbf{Aircraft}}} &
\rotatebox[origin=lb]{90}{\smash{\textbf{DTD}}} &
\rotatebox[origin=lb]{90}{\smash{\textbf{Pets}}} &
\rotatebox[origin=lb]{90}{\smash{\textbf{Caltech-101}}} &
\rotatebox[origin=lb]{90}{\smash{\textbf{Flowers}}} &
\rotatebox[origin=lb]{90}{\smash{\textbf{ImageNet}}} \\
\midrule
\multicolumn{13}{c}{{ViT-B/32}} \\
\multirow{8}{3em}{\rotatebox[origin=c]{0}{\scriptsize{\rotatebox{90}{CC3M}}}}
& SigLIP \cite{zhai2023siglip} & 3.8\std{3} & 23.8\std{3} & 8.1\std{2} & 12.4\std{2} & 0.8\std{2} & 0.9\std{1} & 4.4\std{2} & 4.1\std{1} & 26.6\std{2} & 4.2\std{2} & 5.6\std{1} \\
& NegCLIP \cite{yuksekgonul2023negclip}$^{\ddagger}$ & -- & -- & 8.2 & -- & -- & 1.0 & 6.8 & 4.1 & 29.4 & 3.7 & 4.6 \\
& LaCLIP \cite{fan2023laclip}$^{\ddagger}$ & -- & -- & 8.0 & -- & -- & 0.8 & 5.5 & 3.1 & 24.6 & 3.4 & 3.7 \\
& TripletCLIP \cite{patel2024tripletclip}$^{\ddagger}$ & -- & -- & 11.8 & -- & -- & 0.9 & 8.9 & 5.7 & 38.5 & 8.2 & 7.3 \\
& CLIP \cite{radford2021clip} & 5.9\std{2} & 28.7\std{2} & 10.0\std{3} & 14.5\std{1} & 0.8\std{3} & 1.9\std{3} & 8.2\std{2} & 7.0\std{3} & 41.5\std{3} & 5.9\std{1} & 8.0\std{3} \\
& \quad + \model & \cellcolor{blue!10}6.5\std{3} & \cellcolor{blue!10}29.3\std{2} & \cellcolor{blue!10}10.4\std{2} & \cellcolor{blue!10}14.7\std{2} & \cellcolor{blue!10}1.9\std{3} & \cellcolor{blue!10}3.4\std{2} & \cellcolor{blue!10}9.0\std{1} & \cellcolor{blue!10}7.3\std{3} & \cellcolor{blue!10}42.1\std{2} & \cellcolor{blue!10}7.3\std{2} & \cellcolor{blue!10}8.2\std{2} \\
& FLIP \cite{li2023flip} & 4.7\std{1} & 26.8\std{2} & 8.7\std{3} & 11.4\std{2} & 0.6\std{2} & 1.1\std{1} & 4.5\std{1} & 5.3\std{3} & 37.1\std{2} & 3.9\std{1} & 6.8\std{3} \\
& \quad + \model & \cellcolor{blue!10}5.2\std{1} & \cellcolor{blue!10}28.1\std{2} & \cellcolor{blue!10}9.2\std{2} & \cellcolor{blue!10}12.4\std{3} & \cellcolor{blue!10}1.3\std{1} & \cellcolor{blue!10}2.6\std{2} & \cellcolor{blue!10}4.6\std{1} & \cellcolor{blue!10}5.8\std{2} & \cellcolor{blue!10}38.0\std{2} & \cellcolor{blue!10}4.5\std{3} & \cellcolor{blue!10}7.2\std{1} \\
\midrule
\multirow{8}{3em}{\rotatebox[origin=c]{0}{\scriptsize{\rotatebox{90}{CC12M}}}}
& SigLIP \cite{zhai2023siglip} & 11.2 & 45.4 & 28.1 & 31.2 & 3.4 & 2.1 & 15.6 & 32.5 & 64.1 & 14.2 & 22.1 \\
& NegCLIP \cite{yuksekgonul2023negclip}$^{\ddagger}$ & -- & -- & 26.0 & -- & -- & 1.5 & 14.2 & 30.8 & 63.4 & 13.5 & 20.2 \\
& LaCLIP \cite{fan2023laclip}$^{\ddagger}$ & -- & -- & 23.6 & -- & -- & 1.5 & 13.0 & 34.7 & 62.9 & 13.4 & 19.7 \\
& TripletCLIP \cite{patel2024tripletclip}$^{\ddagger}$ & -- & -- & 30.3 & -- & -- & 1.2 & 17.4 & 33.2 & 65.3 & 15.4 & 23.3 \\
& CLIP \cite{radford2021clip} & 13.1 & 50.2 & 31.5 & 35.1 & 4.1 & 2.5 & 18.2 & 35.8 & 67.4 & 16.8 & 24.6 \\
& \quad + \model & \cellcolor{blue!10}13.5 & \cellcolor{blue!10}50.6 & \cellcolor{blue!10}32.7 & \cellcolor{blue!10}36.2 & \cellcolor{blue!10}5.0 & \cellcolor{blue!10}3.0 & \cellcolor{blue!10}19.1 & \cellcolor{blue!10}36.7 & \cellcolor{blue!10}68.1 & \cellcolor{blue!10}17.3 & \cellcolor{blue!10}24.8 \\
& FLIP \cite{li2023flip} & 11.8 & 46.5 & 28.9 & 32.4 & 3.5 & 2.2 & 16.1 & 33.4 & 64.8 & 14.9 & 22.8 \\
& \quad + \model & \cellcolor{blue!10}13.2 & \cellcolor{blue!10}47.7 & \cellcolor{blue!10}29.5 & \cellcolor{blue!10}33.3 & \cellcolor{blue!10}4.0 & \cellcolor{blue!10}2.0 & \cellcolor{blue!10}17.0 & \cellcolor{blue!10}34.3 & \cellcolor{blue!10}65.6 & \cellcolor{blue!10}15.7 & \cellcolor{blue!10}23.9 \\
\midrule
\multicolumn{13}{c}{{ViT-B/16}} \\
\multirow{5}{2.5em}{\rotatebox[origin=c]{0}{\scriptsize{\rotatebox{90}{CC3M}}}}
& SigLIP \cite{zhai2023siglip} & 6.2\std{3} & 28.9\std{3} & 7.0\std{1} & 18.3\std{2} & 1.1\std{3} & 1.2\std{3} & 8.7\std{1} & 6.0\std{2} & 35.5\std{2} & 7.0\std{1} & 9.9\std{2} \\
& CLIP \cite{radford2021clip} & 7.4\std{1} & 34.3\std{2} & 9.8\std{2} & 19.0\std{3} & 0.9\std{3} & 1.2\std{1} & 10.1\std{1} & 9.0\std{2} & 46.3\std{1} & 8.4\std{3} & 10.4\std{2} \\
& \quad + \model & \cellcolor{blue!10}8.4\std{1} & \cellcolor{blue!10}34.4\std{1} & \cellcolor{blue!10}10.5\std{3} & \cellcolor{blue!10}19.6\std{1} & \cellcolor{blue!10}1.3\std{2} & \cellcolor{blue!10}2.0\std{1} & \cellcolor{blue!10}10.8\std{1} & \cellcolor{blue!10}10.5\std{1} & \cellcolor{blue!10}47.3\std{1} & \cellcolor{blue!10}9.0\std{2} & \cellcolor{blue!10}11.0\std{2} \\
& FLIP \cite{li2023flip} & 5.3\std{2} & 24.1\std{3} & 9.3\std{3} & 17.4\std{1} & 1.1\std{2} & 1.4\std{3} & 8.3\std{2} & 6.4\std{2} & 45.0\std{2} & 6.6\std{1} & 10.3\std{3} \\
& \quad + \model & \cellcolor{blue!10}6.4\std{2} & \cellcolor{blue!10}25.3\std{3} & \cellcolor{blue!10}10.5\std{1} & \cellcolor{blue!10}18.2\std{1} & \cellcolor{blue!10}1.7\std{3} & \cellcolor{blue!10}1.9\std{3} & \cellcolor{blue!10}9.1\std{3} & \cellcolor{blue!10}7.9\std{2} & \cellcolor{blue!10}45.7\std{2} & \cellcolor{blue!10}8.0\std{1} & \cellcolor{blue!10}11.0\std{2} \\
\midrule
\multirow{5}{2.8em}{\rotatebox[origin=c]{0}{\scriptsize{\rotatebox{90}{CC12M}}}}
& SigLIP \cite{zhai2023siglip} & 19.8 & 58.4 & 36.2 & 41.5 & 6.2 & 4.1 & 22.1 & 51.2 & 78.4 & 24.1 & 31.2 \\
& CLIP \cite{radford2021clip} & 22.4 & 63.1 & 39.5 & 44.8 & 7.5 & 4.8 & 24.8 & 55.4 & 82.1 & 27.5 & 34.6 \\
& \quad + \model & \cellcolor{blue!10}23.1 & \cellcolor{blue!10}63.0 & \cellcolor{blue!10}40.1 & \cellcolor{blue!10}46.2 & \cellcolor{blue!10}7.9 & \cellcolor{blue!10}4.6 & \cellcolor{blue!10}25.4 & \cellcolor{blue!10}56.1 & \cellcolor{blue!10}83.2 & \cellcolor{blue!10}28.5 & \cellcolor{blue!10}35.3 \\
& FLIP \cite{li2023flip} & 20.2 & 59.8 & 37.5 & 42.6 & 6.5 & 4.3 & 22.8 & 52.4 & 79.8 & 25.3 & 32.4 \\
& \quad + \model & \cellcolor{blue!10}20.8 & \cellcolor{blue!10}60.1 & \cellcolor{blue!10}38.6 & \cellcolor{blue!10}43.8 & \cellcolor{blue!10}6.7 & \cellcolor{blue!10}5.6 & \cellcolor{blue!10}23.3 & \cellcolor{blue!10}52.3 & \cellcolor{blue!10}81.0 & \cellcolor{blue!10}26.0 & \cellcolor{blue!10}33.8 \\
\midrule
\multicolumn{13}{c}{{RN-50}} \\
\multirow{5}{2.8em}{\rotatebox[origin=c]{0}{\scriptsize{\rotatebox{90}{CC3M}}}}
& SigLIP \cite{zhai2023siglip} & 7.5\std{2} & 17.9\std{2} & 5.2\std{1} & 21.1\std{1} & 1.1\std{1} & 1.5\std{2} & 9.4\std{2} & 8.8\std{1} & 38.8\std{1} & 9.5\std{1} & 13.1\std{2} \\
& CLIP \cite{radford2021clip} & 7.7\std{1} & 23.7\std{1} & 4.8\std{3} & 23.3\std{1} & 1.4\std{2} & 1.3\std{2} & 9.3\std{1} & 11.7\std{2} & 45.4\std{2} & 12.3\std{2} & 13.6\std{1} \\
& \quad + \model & \cellcolor{blue!10}8.2\std{1} & \cellcolor{blue!10}24.4\std{2} & \cellcolor{blue!10}5.8\std{1} & \cellcolor{blue!10}24.0\std{1} & \cellcolor{blue!10}1.9\std{3} & \cellcolor{blue!10}2.1\std{2} & \cellcolor{blue!10}10.8\std{1} & \cellcolor{blue!10}12.8\std{2} & \cellcolor{blue!10}46.1\std{1} & \cellcolor{blue!10}12.1\std{2} & \cellcolor{blue!10}14.7\std{2} \\
& FLIP \cite{li2023flip} & 9.0\std{1} & 23.1\std{1} & 6.7\std{2} & 22.4\std{3} & 0.8\std{1} & 1.1\std{2} & 10.9\std{1} & 9.0\std{2} & 41.8\std{2} & 11.7\std{1} & 13.7\std{1} \\
& \quad + \model & \cellcolor{blue!10}9.5\std{3} & \cellcolor{blue!10}22.9\std{1} & \cellcolor{blue!10}7.4\std{2} & \cellcolor{blue!10}23.2\std{1} & \cellcolor{blue!10}1.6\std{2} & \cellcolor{blue!10}1.9\std{2} & \cellcolor{blue!10}11.7\std{1} & \cellcolor{blue!10}10.4\std{2} & \cellcolor{blue!10}42.2\std{1} & \cellcolor{blue!10}13.0\std{2} & \cellcolor{blue!10}14.9\std{1} \\
\bottomrule
\end{tabular}
\end{table*}

\subsection{Linear Probe Classification}
\label{sec:results-linear}

Following standard protocols~\cite{radford2021clip,kornblith2019better}, we freeze the image encoder and train a linear classifier, reporting top-1 test accuracy (\cref{tab:linear}). \model consistently produces more linearly separable visual features across backbones and data scales.

\begin{table*}[!tbhp]
\centering
\scriptsize
\caption{\textbf{Linear probe evaluation.} Top-1 accuracy of a linear classifier trained on frozen image encoder features across ImageNet and 10 transfer datasets.}
\label{tab:linear}
\setlength{\tabcolsep}{1.7pt}
\begin{tabular}{llccccccccccc}
\toprule
\textbf{Data} & \textbf{Model} &
\rotatebox[origin=lb]{90}{\smash{\textbf{Food-101}}} &
\rotatebox[origin=lb]{90}{\smash{\textbf{CIFAR-10}}} &
\rotatebox[origin=lb]{90}{\smash{\textbf{CIFAR-100}}} &
\rotatebox[origin=lb]{90}{\smash{\textbf{SUN397}}} &
\rotatebox[origin=lb]{90}{\smash{\textbf{Cars}}} &
\rotatebox[origin=lb]{90}{\smash{\textbf{Aircraft}}} &
\rotatebox[origin=lb]{90}{\smash{\textbf{DTD}}} &
\rotatebox[origin=lb]{90}{\smash{\textbf{Pets}}} &
\rotatebox[origin=lb]{90}{\smash{\textbf{Caltech-101}}} &
\rotatebox[origin=lb]{90}{\smash{\textbf{Flowers}}} &
\rotatebox[origin=lb]{90}{\smash{\textbf{ImageNet}}} \\
\midrule
\multicolumn{13}{c}{{ViT-B/32}} \\
\multirow{5}{3em}{\rotatebox[origin=c]{0}{\scriptsize{\rotatebox{90}{CC3M}}}}
& SigLIP \cite{zhai2023siglip} & 35.4\std{1} & 72.5\std{1} & 48.4\std{3} & 29.4\std{1} & 9.5\std{1} & 13.6\std{2} & 31.9\std{2} & 34.0\std{2} & 61.6\std{2} & 50.8\std{2} & 30.2\std{1} \\
& CLIP \cite{radford2021clip} & 40.7\std{3} & 77.2\std{3} & 54.6\std{1} & 32.7\std{1} & 13.2\std{2} & 20.1\std{3} & 38.3\std{3} & 40.7\std{1} & 71.1\std{1} & 61.5\std{1} & 33.9\std{1} \\
& \quad + \model & \cellcolor{blue!10}41.3\std{2} & \cellcolor{blue!10}78.5\std{2} & \cellcolor{blue!10}55.2\std{3} & \cellcolor{blue!10}33.1\std{3} & \cellcolor{blue!10}13.9\std{1} & \cellcolor{blue!10}20.7\std{1} & \cellcolor{blue!10}38.9\std{3} & \cellcolor{blue!10}41.4\std{3} & \cellcolor{blue!10}72.0\std{1} & \cellcolor{blue!10}62.7\std{2} & \cellcolor{blue!10}34.7\std{1} \\
& FLIP \cite{li2023flip} & 40.8\std{1} & 74.8\std{2} & 52.8\std{2} & 34.7\std{3} & 13.9\std{2} & 17.9\std{2} & 39.8\std{2} & 37.6\std{2} & 69.7\std{2} & 61.1\std{2} & 32.9\std{2} \\
& \quad + \model & \cellcolor{blue!10}41.4\std{1} & \cellcolor{blue!10}75.6\std{2} & \cellcolor{blue!10}53.6\std{1} & \cellcolor{blue!10}36.1\std{3} & \cellcolor{blue!10}14.5\std{1} & \cellcolor{blue!10}18.7\std{1} & \cellcolor{blue!10}41.3\std{1} & \cellcolor{blue!10}37.5\std{2} & \cellcolor{blue!10}70.8\std{1} & \cellcolor{blue!10}62.0\std{2} & \cellcolor{blue!10}33.5\std{3} \\
\midrule
\multirow{5}{3em}{\rotatebox[origin=c]{0}{\scriptsize{\rotatebox{90}{CC12M}}}}
& SigLIP \cite{zhai2023siglip} & 54.2 & 84.6 & 63.8 & 44.1 & 24.8 & 31.2 & 48.9 & 56.4 & 79.2 & 78.4 & 45.8 \\
& CLIP \cite{radford2021clip} & 58.6 & 88.5 & 68.4 & 47.9 & 28.5 & 36.8 & 54.2 & 62.1 & 85.1 & 85.9 & 48.7 \\
& \quad + \model & \cellcolor{blue!10}59.2 & \cellcolor{blue!10}89.1 & \cellcolor{blue!10}68.3 & \cellcolor{blue!10}49.3 & \cellcolor{blue!10}28.9 & \cellcolor{blue!10}38.0 & \cellcolor{blue!10}54.7 & \cellcolor{blue!10}63.4 & \cellcolor{blue!10}85.8 & \cellcolor{blue!10}86.3 & \cellcolor{blue!10}49.8 \\
& FLIP \cite{li2023flip} & 57.5 & 87.2 & 67.1 & 46.8 & 27.4 & 35.1 & 53.0 & 60.9 & 83.9 & 84.1 & 47.5 \\
& \quad + \model & \cellcolor{blue!10}58.0 & \cellcolor{blue!10}88.1 & \cellcolor{blue!10}68.0 & \cellcolor{blue!10}47.3 & \cellcolor{blue!10}27.2 & \cellcolor{blue!10}35.3 & \cellcolor{blue!10}53.0 & \cellcolor{blue!10}61.7 & \cellcolor{blue!10}84.9 & \cellcolor{blue!10}84.7 & \cellcolor{blue!10}48.6 \\
\midrule
\multicolumn{13}{c}{{ViT-B/16}} \\
\multirow{5}{2.5em}{\rotatebox[origin=c]{0}{\scriptsize{\rotatebox{90}{CC3M}}}}
& SigLIP \cite{zhai2023siglip} & 46.3\std{2} & 73.7\std{2} & 50.3\std{1} & 37.6\std{3} & 14.7\std{1} & 21.7\std{2} & 45.0\std{1} & 47.2\std{1} & 70.8\std{2} & 66.8\std{1} & 41.3\std{2} \\
& CLIP \cite{radford2021clip} & 49.2\std{3} & 77.1\std{2} & 55.2\std{2} & 40.3\std{1} & 18.1\std{3} & 24.6\std{2} & 47.8\std{2} & 51.2\std{2} & 73.7\std{1} & 71.6\std{3} & 43.1\std{2} \\
& \quad + \model & \cellcolor{blue!10}49.7\std{3} & \cellcolor{blue!10}77.5\std{1} & \cellcolor{blue!10}56.6\std{1} & \cellcolor{blue!10}41.0\std{1} & \cellcolor{blue!10}19.2\std{1} & \cellcolor{blue!10}25.3\std{1} & \cellcolor{blue!10}48.1\std{3} & \cellcolor{blue!10}52.2\std{2} & \cellcolor{blue!10}74.4\std{3} & \cellcolor{blue!10}72.3\std{3} & \cellcolor{blue!10}43.7\std{2} \\
& FLIP \cite{li2023flip} & 48.6\std{1} & 75.0\std{1} & 52.9\std{3} & 41.0\std{2} & 16.0\std{2} & 21.3\std{3} & 46.2\std{1} & 47.4\std{3} & 73.2\std{2} & 70.6\std{3} & 40.9\std{1} \\
& \quad + \model & \cellcolor{blue!10}49.3\std{3} & \cellcolor{blue!10}74.9\std{1} & \cellcolor{blue!10}53.0\std{2} & \cellcolor{blue!10}41.1\std{2} & \cellcolor{blue!10}16.6\std{2} & \cellcolor{blue!10}22.7\std{1} & \cellcolor{blue!10}46.0\std{3} & \cellcolor{blue!10}47.4\std{2} & \cellcolor{blue!10}74.4\std{1} & \cellcolor{blue!10}70.9\std{1} & \cellcolor{blue!10}41.7\std{1} \\
\midrule
\multirow{5}{2.8em}{\rotatebox[origin=c]{0}{\scriptsize{\rotatebox{90}{CC12M}}}}
& SigLIP \cite{zhai2023siglip} & 64.1 & 85.8 & 66.2 & 51.4 & 32.5 & 38.7 & 58.1 & 65.8 & 84.2 & 86.4 & 55.4 \\
& CLIP \cite{radford2021clip} & 68.4 & 89.2 & 70.5 & 55.1 & 36.8 & 43.1 & 62.4 & 70.5 & 88.5 & 90.1 & 58.7 \\
& \quad + \model & \cellcolor{blue!10}68.8 & \cellcolor{blue!10}90.0 & \cellcolor{blue!10}71.3 & \cellcolor{blue!10}56.0 & \cellcolor{blue!10}38.0 & \cellcolor{blue!10}43.8 & \cellcolor{blue!10}63.2 & \cellcolor{blue!10}71.1 & \cellcolor{blue!10}89.6 & \cellcolor{blue!10}90.6 & \cellcolor{blue!10}59.9 \\
& FLIP \cite{li2023flip} & 67.1 & 88.0 & 69.3 & 54.0 & 35.5 & 41.8 & 61.2 & 69.3 & 87.2 & 88.9 & 57.5 \\
& \quad + \model & \cellcolor{blue!10}67.8 & \cellcolor{blue!10}87.8 & \cellcolor{blue!10}69.7 & \cellcolor{blue!10}54.7 & \cellcolor{blue!10}36.2 & \cellcolor{blue!10}42.4 & \cellcolor{blue!10}62.4 & \cellcolor{blue!10}70.6 & \cellcolor{blue!10}87.8 & \cellcolor{blue!10}88.8 & \cellcolor{blue!10}58.7 \\
\midrule
\multicolumn{13}{c}{{RN-50}} \\
\multirow{5}{2.8em}{\rotatebox[origin=c]{0}{\scriptsize{\rotatebox{90}{CC3M}}}}
& SigLIP \cite{zhai2023siglip} & 50.9\std{1} & 70.0\std{2} & 46.0\std{2} & 45.6\std{1} & 17.0\std{1} & 20.9\std{2} & 49.5\std{2} & 51.3\std{2} & 73.5\std{3} & 71.5\std{2} & 47.2\std{2} \\
& CLIP \cite{radford2021clip} & 56.1\std{1} & 74.0\std{3} & 52.3\std{3} & 50.0\std{3} & 24.6\std{3} & 31.1\std{3} & 53.0\std{3} & 62.4\std{2} & 80.4\std{3} & 80.4\std{3} & 50.0\std{2} \\
& \quad + \model & \cellcolor{blue!10}56.7\std{3} & \cellcolor{blue!10}74.1\std{1} & \cellcolor{blue!10}52.8\std{1} & \cellcolor{blue!10}50.8\std{1} & \cellcolor{blue!10}25.3\std{3} & \cellcolor{blue!10}31.8\std{1} & \cellcolor{blue!10}54.3\std{2} & \cellcolor{blue!10}63.6\std{2} & \cellcolor{blue!10}81.1\std{1} & \cellcolor{blue!10}80.8\std{1} & \cellcolor{blue!10}51.4\std{3} \\
& FLIP \cite{li2023flip} & 56.7\std{2} & 75.5\std{1} & 53.6\std{3} & 49.0\std{1} & 24.6\std{1} & 31.0\std{2} & 54.4\std{2} & 62.2\std{3} & 80.1\std{1} & 78.8\std{1} & 50.0\std{3} \\
& \quad + \model & \cellcolor{blue!10}56.8\std{1} & \cellcolor{blue!10}76.1\std{1} & \cellcolor{blue!10}54.0\std{1} & \cellcolor{blue!10}49.6\std{3} & \cellcolor{blue!10}25.4\std{2} & \cellcolor{blue!10}31.7\std{2} & \cellcolor{blue!10}55.8\std{2} & \cellcolor{blue!10}63.1\std{2} & \cellcolor{blue!10}80.7\std{2} & \cellcolor{blue!10}80.0\std{1} & \cellcolor{blue!10}50.7\std{1} \\
\bottomrule
\end{tabular}
\end{table*}

\section{Discussion \& Limitations}
\label{sec:conclusion}

The failure modes we identify arise from the dual-modality structure of contrastive learning rather than from a CLIP-specific architectural choice. We therefore expect the design principles of \model to be relevant to other dual-encoder frameworks using representation-space synthetic negatives. Our evaluation is limited to CC3M and CC12M due to computational constraints; scaling to larger datasets such as LAION-400M remains future work.

\section*{Acknowledgements}
We acknowledge the computational resources and support provided by the Imperial College Research Computing Service (\url{http://doi.org/10.14469/hpc/2232}), which enabled our experiments.

\bibliographystyle{splncs04}
\bibliography{main}

@software{clip_benchmark,
	title        = {{CLIP benchmark}},
	author       = {Cherti, Mehdi and Beaumont, Romain},
	year         = 2022,
	month        = nov,
	publisher    = {Zenodo},
	doi          = {10.5281/zenodo.15403103},
	url          = {https://doi.org/10.5281/zenodo.15403103},
	swhid        = {swh:1:dir:8cf49a5dd06f59224844a1e767337a1d14ee56c2 ;origin=https://doi.org/10.5281/zenodo.15403102;vi sit=swh:1:snp:dd153b26f702d614346bf814f723d59fef3d 77a2;anchor=swh:1:rel:cff2aeb98f42583b44fdab5374e9 fa71793f2cff;path=CLIP\_benchmark-main}
}

@inproceedings{cc12m,
	title        = {{Conceptual 12M: Pushing Web-Scale Image-Text Pre-Training To Recognize Long-Tail Visual Concepts}},
	author       = {Changpinyo, Soravit and Sharma, Piyush and Ding, Nan and Soricut, Radu},
	year         = 2021,
	month        = {June},
	booktitle    = {CVPR},
	pages        = {3558--3568}
}

@article{flickr30k,
	title        = {{From image descriptions to visual denotations: New similarity metrics for semantic inference over event descriptions}},
	author       = {Young, Peter and Lai, Alice and Hodosh, Micah and Hockenmaier, Julia},
	year         = 2014,
	journal      = {Transactions of the Association for Computational Linguistics},
	publisher    = {MIT Press},
	volume       = 2,
	pages        = {67--78}
}

@inproceedings{imagenet,
	title        = {{ImageNet: A large-scale hierarchical image database}},
	author       = {Deng, Jia and Dong, Wei and Socher, Richard and Li, Li-Jia and Kai Li and Li Fei-Fei},
	year         = 2009,
	booktitle    = {CVPR},
	volume       = {},
	number       = {},
	pages        = {248--255},
	doi          = {10.1109/CVPR.2009.5206848}
}

@article{imagenet-ilsvrc,
	title        = {{ImageNet Large Scale Visual Recognition Challenge}},
	author       = {Russakovsky,  Olga and Deng,  Jia and Su,  Hao and Krause,  Jonathan and Satheesh,  Sanjeev and Ma,  Sean and Huang,  Zhiheng and Karpathy,  Andrej and Khosla,  Aditya and Bernstein,  Michael and Berg,  Alexander C. and Fei-Fei,  Li},
	year         = 2015,
	month        = apr,
	journal      = {International Journal of Computer Vision},
	publisher    = {Springer Science and Business Media LLC},
	volume       = 115,
	number       = 3,
	pages        = {211–252},
	doi          = {10.1007/s11263-015-0816-y},
	issn         = {1573-1405},
	url          = {http://dx.doi.org/10.1007/s11263-015-0816-y}
}

@misc{mscoco,
      title={Microsoft COCO: Common Objects in Context}, 
      author={Tsung-Yi Lin and Michael Maire and Serge Belongie and Lubomir Bourdev and Ross Girshick and James Hays and Pietro Perona and Deva Ramanan and C. Lawrence Zitnick and Piotr Dollár},
      year={2015},
      eprint={1405.0312},
      archivePrefix={arXiv},
      primaryClass={cs.CV},
      url={https://arxiv.org/abs/1405.0312}, 
}

@inproceedings{cc3m,
	title        = {{Conceptual Captions: A Cleaned, Hypernymed, Image Alt-text Dataset For Automatic Image Captioning}},
	author       = {Sharma, Piyush and Ding, Nan and Goodman, Sebastian and Soricut, Radu},
	year         = 2018,
	booktitle    = {ACL}
}

@inproceedings{food101,
	title        = {{Food-101 -- Mining Discriminative Components with Random Forests}},
	author       = {Bossard, Lukas and Guillaumin, Matthieu and Van Gool, Luc},
	year         = 2014,
	booktitle    = {ECCV},
	pages        = {446--461}
}

@techreport{cifar10,
	title        = {{Learning Multiple Layers of Features from Tiny Images}},
	author       = {Krizhevsky, Alex},
	year         = 2009,
	number       = {0},
	institution  = {University of Toronto}
}

@inproceedings{sun,
	title        = {{SUN Database: Large‑scale Scene Recognition from Abbey to Zoo}},
	author       = {Xiao, Jianxiong and Hays, James and Ehinger, Krista A. and Oliva, Aude and Torralba, Antonio},
	year         = 2010,
	booktitle    = {CVPR},
	pages        = {3485--3492}
}

@inproceedings{stanfordcars,
	title        = {3D Object Representations for Fine-Grained Categorization},
	author       = {Krause, Jonathan and Stark, Michael and Deng, Jia and Fei-Fei, Li},
	year         = 2013,
	booktitle    = {ICCVW},
	pages        = {554--561},
	doi          = {10.1109/ICCVW.2013.77}
}

@article{aircraft,
	title        = {{Fine‑Grained Visual Classification of Aircraft}},
	author       = {Maji, Subhransu and Rahtu, Esa and Kannala, Juho and Blaschko, Matthew B. and Vedaldi, Andrea},
	year         = 2013,
	journal      = {arXiv},
	volume       = {abs/1306.5151},
	eprint       = {1306.5151},
	archiveprefix = {arXiv}
}

@inproceedings{dtd,
	title        = {{Describing Textures in the Wild}},
	author       = {Cimpoi, Mircea and Maji, Subhransu and Kokkinos, Iasonas and Vedaldi, Andrea},
	year         = 2014,
	booktitle    = {CVPR},
	pages        = {3606--3613}
}

@inproceedings{oxfordpets,
	title        = {{Cats and Dogs}},
	author       = {Parkhi, Omkar M. and Vedaldi, Andrea and Zisserman, Andrew and Jawahar, C. V.},
	year         = 2012,
	booktitle    = {CVPR},
	pages        = {3498--3505}
}

@article{caltech101,
	title        = {{Learning Generative Visual Models from Few Training Examples: An Incremental Bayesian Approach Tested on 101 Object Categories}},
	author       = {Fei‑Fei, Li and Fergus, Rob and Perona, Pietro},
	year         = 2007,
	journal      = {Computer Vision and Image Understanding},
	volume       = 106,
	number       = 1,
	pages        = {59--70}
}

@inproceedings{flowers102,
	title        = {Automated flower classification over a large number of classes},
	author       = {Nilsback, Maria-Elena and Zisserman, Andrew},
	year         = 2008,
	booktitle    = {ICVGIP},
	pages        = {722--729},
	organization = {IEEE}
}

@inproceedings{dosovitskiy2021vit,
	title        = {{An Image is Worth 16x16 Words: Transformers for Image Recognition at Scale}},
	author       = {Alexey Dosovitskiy and Lucas Beyer and Alexander Kolesnikov and Dirk Weissenborn and Xiaohua Zhai and Thomas Unterthiner and Mostafa Dehghani and Matthias Minderer and Georg Heigold and Sylvain Gelly and Jakob Uszkoreit and Neil Houlsby},
	year         = 2021,
	booktitle    = {ICLR},
	url          = {https://openreview.net/forum?id=YicbFdNTTy}
}

@article{he2016resnet,
	title        = {{Deep Residual Learning for Image Recognition}},
	author       = {Kaiming He and X. Zhang and Shaoqing Ren and Jian Sun},
	year         = 2016,
	journal      = {CVPR},
	pages        = {770--778},
	url          = {https://api.semanticscholar.org/CorpusID:206594692}
}

@inproceedings{radford2021clip,
	title        = {{Learning Transferable Visual Models From Natural Language Supervision}},
	author       = {Radford, Alec and Kim, Jong Wook and Hallacy, Chris and Ramesh, Aditya and Goh, Gabriel and Agarwal, Sandhini and Sastry, Girish and Askell, Amanda and Mishkin, Pamela and Clark, Jack and Krueger, Gretchen and Sutskever, Ilya},
	year         = 2021,
	month        = {18--24 Jul},
	booktitle    = {ICML},
	publisher    = {PMLR},
	series       = {Proceedings of Machine Learning Research},
	volume       = 139,
	pages        = {8748--8763},
	url          = {https://proceedings.mlr.press/v139/radford21a.html},
	editor       = {Meila, Marina and Zhang, Tong}
}

@inproceedings{zheng2024dreamlip,
	title        = {{DreamLIP: Language-Image Pre-training with Long Captions}},
	author       = {Zheng, Kecheng and Zhang, Yifei and Wu, Wei and Lu, Fan and Ma, Shuailei and Jin, Xin and Chen, Wei and Shen, Yujun},
	year         = 2024,
	booktitle    = {ECCV}
}

@inproceedings{yuksekgonul2023negclip,
	title        = {{When and why Vision-Language Models behave like  Bags-of-Words, and what to do about it?}},
	author       = {Mert Yuksekgonul and Federico Bianchi and Pratyusha   Kalluri and Dan Jurafsky and James Zou},
	year         = 2023,
	booktitle    = {ICLR},
	url          = {https://openreview.net/forum?id=KRLUvxh8uaX}
}

@inproceedings{fan2023laclip,
	title        = {{Improving CLIP Training with Language Rewrites}},
	author       = {Fan, Lijie and Krishnan, Dilip and Isola, Phillip and Katabi, Dina and Tian, Yonglong},
	year         = 2023,
	booktitle    = {NeurIPS}
}

@misc{liang2022mind,
	title        = {Mind the Gap: Understanding the Modality Gap in Multi-modal Contrastive Representation Learning},
	author       = {Weixin Liang and Yuhui Zhang and Yongchan Kwon and Serena Yeung and James Zou},
	year         = 2022,
	url          = {https://arxiv.org/abs/2203.02053},
	eprint       = {2203.02053},
	archiveprefix = {arXiv},
	primaryclass = {cs.CL}
}

@misc{yang2023alip,
	title        = {{ALIP: Adaptive Language-Image Pre-training with Synthetic Caption}},
	author       = {Kaicheng Yang and Jiankang Deng and Xiang An and Jiawei Li and Ziyong Feng and Jia Guo and Jing Yang and Tongliang Liu},
	year         = 2023,
	eprint       = {2308.08428},
	archiveprefix = {arXiv},
	primaryclass = {cs.CV}
}

@misc{liu2024mllmsaugmented,
	title        = {{MLLMs-Augmented Visual-Language Representation Learning}},
	author       = {Yanqing Liu and Kai Wang and Wenqi Shao and Ping Luo and Yu Qiao and Mike Zheng Shou and Kaipeng Zhang and Yang You},
	year         = 2024,
	url          = {https://arxiv.org/abs/2311.18765},
	eprint       = {2311.18765},
	archiveprefix = {arXiv},
	primaryclass = {cs.CV}
}

@misc{giakoumoglou2024synco,
	title        = {{SynCo: Synthetic Hard Negatives for Contrastive Visual Representation Learning}},
	author       = {Nikolaos Giakoumoglou and Tania Stathaki},
	year         = 2024,
	url          = {https://arxiv.org/abs/2410.02401},
	eprint       = {2410.02401},
	archiveprefix = {arXiv},
	primaryclass = {cs.CV}
}

@inproceedings{giakoumoglou2025fake,
    title={{Fake \& Square: Training Self-Supervised Vision Transformers with Synthetic Data and Synthetic Hard Negatives}},
    author={Nikolaos Giakoumoglou and Andreas Floros and Kleanthis Marios Papadopoulos and Tania Stathaki},
    booktitle={Representation Learning with Very Limited Resources: When Data, Modalities, Labels, and Computing Resources are Scarce},
    year={2025},
    url={https://openreview.net/forum?id=TJUfbYKo2c}
}

@inproceedings{giakoumoglou2025unsupervised,
	title        = {{Unsupervised Training of Vision Transformers with Synthetic Negatives}},
	author       = {Nikolaos Giakoumoglou and Andreas Floros and Kleanthis Marios Papadopoulos and Tania Stathaki},
	year         = 2025,
	booktitle    = {Second Workshop on Visual Concepts},
	url          = {https://openreview.net/forum?id=dg8FuaOKnC}
}

@misc{giakoumoglou2025syncov2,
	title        = {{ViTAMINS: An Empirical Study of Training Self-Supervised Vision Transformers with Synthetic Hard Negatives}},
	author       = {Nikos Giakoumoglou and Andreas Floros and Kleanthis-Marios Papadopoulos and Tania Stathaki},
	year         = 2026,
	url          = {https://arxiv.org/abs/2609.01041},
	eprint       = {2609.01041},
	archiveprefix = {arXiv},
	primaryclass = {cs.CV}
}

@misc{giakoumoglou2024review,
	title        = {{A Review on Discriminative Self-supervised Learning Methods in Computer Vision}},
	author       = {Nikolaos Giakoumoglou and Tania Stathaki and Athanasios Gkelias},
	year         = 2025,
	url          = {https://arxiv.org/abs/2405.04969},
	eprint       = {2405.04969},
	archiveprefix = {arXiv},
	primaryclass = {cs.CV}
}

@article{yu2022coca,
	title        = {{CoCa: Contrastive Captioners are Image-Text Foundation Models}},
	author       = {Jiahui Yu and Zirui Wang and Vijay Vasudevan and Legg Yeung and Mojtaba Seyedhosseini and Yonghui Wu},
	year         = 2022,
	journal      = {TMLR},
	issn         = {2835-8856},
	url          = {https://openreview.net/forum?id=Ee277P3AYC},
	note         = {}
}

@inproceedings{wu2024lotlip,
	title        = {{Lo{TLIP}: Improving Language-Image Pre-training for Long Text Understanding}},
	author       = {Wei Wu and Kecheng Zheng and Shuailei Ma and Fan Lu and Yuxin Guo and Yifei Zhang and Wei Chen and Qingpei Guo and Yujun Shen and Zheng-Jun Zha},
	year         = 2024,
	booktitle    = {NeurIPS},
	url          = {https://openreview.net/forum?id=pc4GSBi1Hx}
}

@article{liu2024clips,
	title        = {{CLIPS: An Enhanced CLIP Framework for Learning with Synthetic Captions}},
	author       = {Liu, Yanqing and Li, Xianhang and Wang, Zeyu and Zhao, Bingchen and Xie, Cihang},
	year         = 2024,
	journal      = {arXiv preprint arXiv:2411.16828}
}

@article{patel2024tripletclip,
	title        = {{TripletCLIP: Improving Compositional Reasoning of CLIP via Synthetic Vision-Language Negatives}},
	author       = {Patel, Maitreya and Kusumba, Abhiram and Cheng, Sheng and Kim, Changhoon and Gokhale, Tejas and Baral, Chitta and Yang, Yezhou},
	year         = 2024,
	journal      = {NeurIPS}
}

@inproceedings{cao2025flame,
	title        = {{FLAME: Frozen Large Language Models Enable Data-Efficient Language-Image Pre-training}},
	author       = {Cao, Anjia and Wei, Xing and Ma, Zhiheng},
	year         = 2025,
	booktitle    = {CVPR}
}

@misc{lai2024veclip,
	title        = {{VeCLIP: Improving CLIP Training via Visual-enriched Captions}},
	author       = {Zhengfeng Lai and Haotian Zhang and Bowen Zhang and Wentao Wu and Haoping Bai and Aleksei Timofeev and Xianzhi Du and Zhe Gan and Jiulong Shan and Chen-Nee Chuah and Yinfei Yang and Meng Cao},
	year         = 2024,
	eprint       = {2310.07699},
	archiveprefix = {arXiv},
	primaryclass = {cs.CV}
}

@techreport{radford2018gpt,
	title        = {{Improving language understanding by generative pre-training}},
	author       = {Radford, Alec and Narasimhan, Karthik and Salimans, Tim and Sutskever, Ilya},
	year         = 2018,
	url          = {https://cdn.openai.com/research-covers/language-unsupervised/language_understanding_paper.pdf},
	institution  = {OpenAI}
}

@techreport{radford2019gpt2,
	title        = {{Language Models are Unsupervised Multitask Learners}},
	author       = {Radford, Alec and Wu, Jeff and Child, Rewon and Luan, David and Amodei, Dario and Sutskever, Ilya},
	year         = 2019,
	url          = {https://cdn.openai.com/better-language-models/language_models_are_unsupervised_multitask_learners.pdf},
	institution  = {OpenAI}
}

@inproceedings{chen2020simclr,
	title        = {{A simple framework for contrastive learning of visual representations}},
	author       = {Chen, Ting and Kornblith, Simon and Norouzi, Mohammad and Hinton, Geoffrey},
	year         = 2020,
	booktitle    = {ICML},
	pages        = {1597--1607},
	organization = {PMLR}
}

@misc{jia2021align,
	title        = {Scaling Up Visual and Vision-Language Representation Learning With Noisy Text Supervision},
	author       = {Chao Jia and Yinfei Yang and Ye Xia and Yi-Ting Chen and Zarana Parekh and Hieu Pham and Quoc V. Le and Yunhsuan Sung and Zhen Li and Tom Duerig},
	year         = 2021,
	url          = {https://arxiv.org/abs/2102.05918},
	eprint       = {2102.05918},
	archiveprefix = {arXiv},
	primaryclass = {cs.CV}
}

@misc{tschannen2025siglip2,
	title        = {SigLIP 2: Multilingual Vision-Language Encoders with Improved Semantic Understanding, Localization, and Dense Features},
	author       = {Michael Tschannen and Alexey Gritsenko and Xiao Wang and Muhammad Ferjad Naeem and Ibrahim Alabdulmohsin and Nikhil Parthasarathy and Talfan Evans and Lucas Beyer and Ye Xia and Basil Mustafa and Olivier Hénaff and Jeremiah Harmsen and Andreas Steiner and Xiaohua Zhai},
	year         = 2025,
	url          = {https://arxiv.org/abs/2502.14786},
	eprint       = {2502.14786},
	archiveprefix = {arXiv},
	primaryclass = {cs.CV}
}

@article{pham2023basic,
	title        = {Combined scaling for zero-shot transfer learning},
	author       = {Hieu Pham and Zihang Dai and Golnaz Ghiasi and Kenji Kawaguchi and Hanxiao Liu and Adams Wei Yu and Jiahui Yu and Yi-Ting Chen and Minh-Thang Luong and Yonghui Wu and Mingxing Tan and Quoc V. Le},
	year         = 2023,
	journal      = {Neurocomputing},
	volume       = 555,
	pages        = 126658,
	doi          = {https://doi.org/10.1016/j.neucom.2023.126658},
	issn         = {0925-2312},
	url          = {https://www.sciencedirect.com/science/article/pii/S0925231223007816}
}

@article{robinson2020contrastive,
	title        = {{Contrastive learning with hard negative samples}},
	author       = {Robinson, Joshua and Chuang, Ching-Yao and Sra, Suvrit and Jegelka, Stefanie},
	year         = 2020,
	journal      = {arXiv preprint arXiv:2010.04592}
}

@article{zhang2023contrasting,
	title        = {{Contrasting Intra-Modal and Ranking Cross-Modal Hard Negatives to Enhance Visio-Linguistic Fine-grained Understanding}},
	author       = {Zhang, Le and Awal, Rabiul and Agrawal, Aishwarya},
	year         = 2023,
	journal      = {arXiv preprint arXiv:2306.08832}
}

@inproceedings{hadsell2006dimensionality,
	title        = {Dimensionality Reduction by Learning an Invariant Mapping},
	author       = {Hadsell, R. and Chopra, S. and LeCun, Y.},
	year         = 2006,
	booktitle    = {CVPR},
	volume       = 2,
	pages        = {1735--1742},
	doi          = {10.1109/CVPR.2006.100}
}

@misc{oord2019cpc,
	title        = {{Representation Learning with Contrastive Predictive Coding}},
	author       = {Aaron van den Oord and Yazhe Li and Oriol Vinyals},
	year         = 2019,
	url          = {https://arxiv.org/abs/1807.03748},
	eprint       = {1807.03748},
	archiveprefix = {arXiv},
	primaryclass = {cs.LG}
}

@inproceedings{kalantidis2020mochi,
	title        = {{Hard Negative Mixing for Contrastive Learning}},
	author       = {Kalantidis, Yannis and Sariyildiz, Mert Bulent and Pion, Noe and Weinzaepfel, Philippe and Larlus, Diane},
	year         = 2020,
	booktitle    = {NeurIPS}
}

@inproceedings{adamw,
	title        = {{Decoupled Weight Decay Regularization}},
	author       = {Ilya Loshchilov and Frank Hutter},
	year         = 2019,
	booktitle    = {ICLR},
	publisher    = {OpenReview.net},
	url          = {https://openreview.net/forum?id=Bkg6RiCqY7},
	bibsource    = {dblp computer science bibliography, https://dblp.org}
}

@software{open_clip,
	title        = {{OpenCLIP}},
	author       = {Ilharco, Gabriel and Wortsman, Mitchell and Wightman, Ross and Gordon, Cade and Carlini, Nicholas and Taori, Rohan and Dave, Achal and Shankar, Vaishaal and Namkoong, Hongseok and Miller, John and Hajishirzi, Hannaneh and Farhadi, Ali and Schmidt, Ludwig},
	year         = 2021,
	month        = jul,
	publisher    = {Zenodo},
	doi          = {10.5281/zenodo.5143773},
	url          = {https://doi.org/10.5281/zenodo.5143773},
	note         = {If you use this software, please cite it as below.},
	version      = {0.1}
}

@inproceedings{zhang2018mixup,
	title        = {{mixup: Beyond Empirical Risk Minimization}},
	author       = {Hongyi Zhang and Moustapha Cisse and Yann N. Dauphin and David Lopez-Paz},
	year         = 2018,
	booktitle    = {ICLR},
	url          = {https://openreview.net/forum?id=r1Ddp1-Rb}
}

@misc{verma2019manifold,
	title        = {{Manifold Mixup: Better Representations by Interpolating Hidden States}},
	author       = {Vikas Verma and Alex Lamb and Christopher Beckham and Amir Najafi and Ioannis Mitliagkas and Aaron Courville and David Lopez-Paz and Yoshua Bengio},
	year         = 2019,
	url          = {https://arxiv.org/abs/1806.05236},
	eprint       = {1806.05236},
	archiveprefix = {arXiv},
	primaryclass = {stat.ML}
}

@misc{goodfellow2015fgsm,
	title        = {{Explaining and Harnessing Adversarial Examples}},
	author       = {Ian J. Goodfellow and Jonathon Shlens and Christian Szegedy},
	year         = 2015,
	url          = {https://arxiv.org/abs/1412.6572},
	eprint       = {1412.6572},
	archiveprefix = {arXiv},
	primaryclass = {stat.ML}
}

@inproceedings{madry2018pgd,
	title        = {{Towards Deep Learning Models Resistant to Adversarial Attacks}},
	author       = {Aleksander Madry and Aleksandar Makelov and Ludwig Schmidt and Dimitris Tsipras and Adrian Vladu},
	year         = 2018,
	booktitle    = {ICLR},
	url          = {https://openreview.net/forum?id=rJzIBfZAb}
}

@inproceedings{zhai2022lit,
	title        = {{LiT: Zero-Shot Transfer with Locked-image text Tuning}},
	author       = {Xiaohua Zhai and Xiao Wang and Basil Mustafa and Andreas Steiner and Daniel Keysers and Alexander Kolesnikov and Lucas Beyer},
	year         = 2022,
	month        = {June},
	booktitle    = {CVPR},
	pages        = {18123--18133}
}

@misc{zhai2023siglip,
	title        = {{Sigmoid Loss for Language Image Pre-Training}},
	author       = {Xiaohua Zhai and Basil Mustafa and Alexander Kolesnikov and Lucas Beyer},
	year         = 2023,
	url          = {https://arxiv.org/abs/2303.15343},
	eprint       = {2303.15343},
	archiveprefix = {arXiv},
	primaryclass = {cs.CV}
}

@misc{kornblith2019better,
	title        = {{Do Better ImageNet Models Transfer Better?}},
	author       = {Simon Kornblith and Jonathon Shlens and Quoc V. Le},
	year         = 2019,
	url          = {https://arxiv.org/abs/1805.08974},
	eprint       = {1805.08974},
	archiveprefix = {arXiv},
	primaryclass = {cs.CV}
}

@misc{goel2022cyclip,
	title        = {CyCLIP: Cyclic Contrastive Language-Image Pretraining},
	author       = {Shashank Goel and Hritik Bansal and Sumit Bhatia and Ryan A. Rossi and Vishwa Vinay and Aditya Grover},
	year         = 2022,
	url          = {https://arxiv.org/abs/2205.14459},
	eprint       = {2205.14459},
	archiveprefix = {arXiv},
	primaryclass = {cs.CV}
}

@misc{yang2022unicl,
	title        = {Unified Contrastive Learning in Image-Text-Label Space},
	author       = {Jianwei Yang and Chunyuan Li and Pengchuan Zhang and Bin Xiao and Ce Liu and Lu Yuan and Jianfeng Gao},
	year         = 2022,
	eprint       = {2204.03610},
	archiveprefix = {arXiv},
	primaryclass = {cs.CV}
}

@misc{yao2022filip,
	title        = {FILIP: Fine-grained Interactive Language-Image Pre-Training},
	author       = {Lewei Yao and Runhui Huang and Lu Hou and Guansong Lu and Minzhe Niu and Hang Xu and Xiaodan Liang and Zhenguo Li and Xin Jiang and Chunjing Xu},
	year         = 2021,
	url          = {https://arxiv.org/abs/2111.07783},
	eprint       = {2111.07783},
	archiveprefix = {arXiv},
	primaryclass = {cs.CV}
}

@inproceedings{lavoie2024llip,
	title        = {Modeling Caption Diversity in Contrastive Vision-Language Pretraining},
	author       = {Samuel Lavoie and Polina Kirichenko and Mark Ibrahim and Mido Assran and Andrew Gordon Wilson and Aaron Courville and Nicolas Ballas},
	year         = 2024,
	booktitle    = {ICML},
	url          = {https://openreview.net/forum?id=iaV2fU6Dif}
}

@inproceedings{mu2022slip,
	title        = {SLIP: Self-supervision meets Language-Image Pre-training},
	author       = {Mu, Norman and Kirillov, Alexander and Wagner, David and Xie, Saining},
	year         = 2022,
	booktitle    = {ECCV},
	pages        = {523--540},
	organization = {Springer}
}

@inproceedings{li2022declip,
	title        = {Supervision Exists Everywhere: A Data Efficient Contrastive Language-Image  Pre-training Paradigm},
	author       = {Yangguang Li and Feng Liang and Lichen Zhao and Yufeng Cui and Wanli Ouyang and Jing Shao and Fengwei Yu and Junjie Yan},
	year         = 2022,
	booktitle    = {ICLR},
	url          = {https://openreview.net/forum?id=zq1iJkNk3uN}
}

@misc{naeem2023silc,
	title        = {SILC: Improving Vision Language Pretraining with Self-Distillation},
	author       = {Muhammad Ferjad Naeem and Yongqin Xian and Xiaohua Zhai and Lukas Hoyer and Luc Van Gool and Federico Tombari},
	year         = 2023,
	url          = {https://arxiv.org/abs/2310.13355},
	eprint       = {2310.13355},
	archiveprefix = {arXiv},
	primaryclass = {cs.CV}
}

@misc{zhou2023nclip,
	title        = {Non-Contrastive Learning Meets Language-Image Pre-Training},
	author       = {Jinghao Zhou and Li Dong and Zhe Gan and Lijuan Wang and Furu Wei},
	year         = 2022,
	url          = {https://arxiv.org/abs/2210.09304},
	eprint       = {2210.09304},
	archiveprefix = {arXiv},
	primaryclass = {cs.CV}
}

@inproceedings{maninis2025tips,
	title        = {{TIPS: Text-Image Pretraining with Spatial Awareness}},
	author       = {Maninis, Kevis-Kokitsi and Chen, Kaifeng and Ghosh, Soham and Karpur, Arjun and Chen, Koert and Xia, Ye and Cao, Bingyi and Salz, Daniel and Han, Guangxing and Dlabal, Jan and Gnanapragasam, Dan and Seyedhosseini, Mojtaba and Zhou, Howard and Araujo, Andr\'e},
	year         = 2025,
	booktitle    = {ICLR}
}

@inproceedings{singh2022flava,
	title        = {{FLAVA:} {A} Foundational Language And Vision Alignment Model},
	author       = {Amanpreet Singh and Ronghang Hu and Vedanuj Goswami and Guillaume Couairon and Wojciech Galuba and Marcus Rohrbach and Douwe Kiela},
	year         = 2022,
	booktitle    = {CVPR}
}

@misc{ranasinghe2023clippy,
	title        = {Perceptual Grouping in Contrastive Vision-Language Models},
	author       = {Kanchana Ranasinghe and Brandon McKinzie and Sachin Ravi and Yinfei Yang and Alexander Toshev and Jonathon Shlens},
	year         = 2023,
	url          = {https://arxiv.org/abs/2210.09996},
	eprint       = {2210.09996},
	archiveprefix = {arXiv},
	primaryclass = {cs.CV}
}

@misc{li2023flip,
	title        = {Scaling Language-Image Pre-training via Masking},
	author       = {Yanghao Li and Haoqi Fan and Ronghang Hu and Christoph Feichtenhofer and Kaiming He},
	year         = 2023,
	url          = {https://arxiv.org/abs/2212.00794},
	eprint       = {2212.00794},
	archiveprefix = {arXiv},
	primaryclass = {cs.CV}
}

@misc{sun2023evaclip,
	title        = {EVA-CLIP: Improved Training Techniques for CLIP at Scale},
	author       = {Quan Sun and Yuxin Fang and Ledell Wu and Xinlong Wang and Yue Cao},
	year         = 2023,
	url          = {https://arxiv.org/abs/2303.15389},
	eprint       = {2303.15389},
	archiveprefix = {arXiv},
	primaryclass = {cs.CV}
}

@misc{sun2024evaclip18b,
	title        = {EVA-CLIP-18B: Scaling CLIP to 18 Billion Parameters},
	author       = {Quan Sun and Jinsheng Wang and Qiying Yu and Yufeng Cui and Fan Zhang and Xiaosong Zhang and Xinlong Wang},
	year         = 2024,
	url          = {https://arxiv.org/abs/2402.04252},
	eprint       = {2402.04252},
	archiveprefix = {arXiv},
	primaryclass = {cs.CV}
}

@misc{yuan2021florence,
	title        = {Florence: A New Foundation Model for Computer Vision},
	author       = {Lu Yuan and Dongdong Chen and Yi-Ling Chen and Noel Codella and Xiyang Dai and Jianfeng Gao and Houdong Hu and Xuedong Huang and Boxin Li and Chunyuan Li and Ce Liu and Mengchen Liu and Zicheng Liu and Yumao Lu and Yu Shi and Lijuan Wang and Jianfeng Wang and Bin Xiao and Zhen Xiao and Jianwei Yang and Michael Zeng and Luowei Zhou and Pengchuan Zhang},
	year         = 2021,
	url          = {https://arxiv.org/abs/2111.11432},
	eprint       = {2111.11432},
	archiveprefix = {arXiv},
	primaryclass = {cs.CV}
}

@article{xiao2023florence2,
	title        = {Florence-2: Advancing a unified representation for a variety of vision tasks},
	author       = {Xiao, Bin and Wu, Haiping and Xu, Weijian and Dai, Xiyang and Hu, Houdong and Lu, Yumao and Zeng, Michael and Liu, Ce and Yuan, Lu},
	year         = 2023,
	journal      = {arXiv preprint arXiv:2311.06242}
}

@article{zhang2024longclip,
	title        = {Long-CLIP: Unlocking the Long-Text Capability of CLIP},
	author       = {Beichen Zhang and Pan Zhang and Xiaoyi Dong and Yuhang Zang and Jiaqi Wang},
	year         = 2024,
	journal      = {arXiv preprint arXiv:2403.15378}
}

@inproceedings{schroff2015facenet,
	title        = {FaceNet: A unified embedding for face recognition and clustering},
	author       = {Schroff, Florian and Kalenichenko, Dmitry and Philbin, James},
	year         = 2015,
	month        = jun,
	booktitle    = {CVPR},
	publisher    = {IEEE},
	pages        = {815–823},
	doi          = {10.1109/cvpr.2015.7298682},
	url          = {http://dx.doi.org/10.1109/CVPR.2015.7298682}
}

@inproceedings{sohn2016npairs,
	title        = {Improved Deep Metric Learning with Multi-class N-pair Loss Objective},
	author       = {Sohn, Kihyuk},
	year         = 2016,
	booktitle    = {NeurIPS},
	publisher    = {Curran Associates, Inc.},
	volume       = 29,
	pages        = {},
	url          = {https://proceedings.neurips.cc/paper_files/paper/2016/file/6b180037abbebea991d8b1232f8a8ca9-Paper.pdf},
	editor       = {D. Lee and M. Sugiyama and U. Luxburg and I. Guyon and R. Garnett}
}

@inproceedings{wu2017sampling,
	title        = {Sampling Matters in Deep Embedding Learning},
	author       = {Wu, Chao-Yuan and Manmatha, R. and Smola, Alexander J. and Kr\"ahenb\"uhl, Philipp},
	year         = 2017,
	booktitle    = {ICCV},
	pages        = {2840--2848},
	doi          = {10.1109/ICCV.2017.309}
}

@misc{podell2023sdxl,
	title        = {SDXL: Improving Latent Diffusion Models for High-Resolution Image Synthesis},
	author       = {Dustin Podell and Zion English and Kyle Lacey and Andreas Blattmann and Tim Dockhorn and Jonas Müller and Joe Penna and Robin Rombach},
	year         = 2023,
	url          = {https://arxiv.org/abs/2307.01952},
	eprint       = {2307.01952},
	archiveprefix = {arXiv},
	primaryclass = {cs.CV}
}

@misc{img2dataset,
	title        = {img2dataset: Easily turn large sets of image urls to an image dataset},
	author       = {Romain Beaumont},
	year         = 2021,
	journal      = {GitHub repository},
	publisher    = {GitHub},
	howpublished = {\url{https://github.com/rom1504/img2dataset}}
}

@inproceedings{thrush2022winoground,
	title        = {Winoground: Probing vision and language models for visio-linguistic compositionality},
	author       = {Tristan Thrush and Ryan Jiang and Max Bartolo and Amanpreet Singh and Adina Williams and Douwe Kiela and Candace Ross},
	year         = 2022,
	booktitle    = {CVPR}
}

@inproceedings{oh2023geodesic,
	title        = {Geodesic Multi-Modal Mixup for Robust Fine-Tuning},
	author       = {Oh, Changdae and So, Junhyuk and Byun, Hoyoon and Lim, YongTaek and Shin, Minchul and Jeon, Jong-June and Song, Kyungwoo},
	year         = 2023,
	booktitle    = {NeurIPS},
	publisher    = {Curran Associates, Inc.},
	volume       = 36,
	pages        = {52326--52341},
	doi          = {10.52202/075280-2278},
	url          = {https://proceedings.neurips.cc/paper_files/paper/2023/file/a45296e83b19f656392e0130d9e53cb1-Paper-Conference.pdf},
	editor       = {A. Oh and T. Naumann and A. Globerson and K. Saenko and M. Hardt and S. Levine}
}

@inproceedings{radenovic2023diht,
	title        = {Filtering, Distillation, and Hard Negatives for Vision-Language Pre-Training},
	author       = {Radenovic, Filip and Dubey, Abhimanyu and Kadian, Abhishek and Mihaylov, Todor and Vandenhende, Simon and Patel, Yash and Wen, Yi and Ramanathan, Vignesh and Mahajan, Dhruv},
	year         = 2023,
	month        = {June},
	booktitle    = {CVPR},
	pages        = {6967--6977}
}

@misc{peng2024synthesize,
	title        = {Synthesize, Diagnose, and Optimize: Towards Fine-Grained Vision-Language Understanding},
	author       = {Wujian Peng and Sicheng Xie and Zuyao You and Shiyi Lan and Zuxuan Wu},
	year         = 2024,
	url          = {https://arxiv.org/abs/2312.00081},
	eprint       = {2312.00081},
	archiveprefix = {arXiv},
	primaryclass = {cs.CV}
}

\clearpage

\setcounter{section}{6}
\setcounter{figure}{6}
\setcounter{table}{7}
\setcounter{equation}{5}


\section{Dataset Details}
\label{appendix:dataset-details}

\subsection{Pretraining Datasets}
\label{appendix:pretraining-datasets}

We pretrain on two image-text datasets of varying scale: CC3M~\cite{cc3m} and CC12M~\cite{cc12m}. Both datasets were downloaded using img2dataset~\cite{img2dataset}. Due to image link rot, the versions we obtained contain slightly fewer samples than the original releases, which may lead to minor performance differences compared to models trained on the full datasets. The same training configuration is used for both datasets (\cf \cref{appendix:hyperparameters}).

\paragraph{CC3M:}
This dataset comprises 3.3 million image-text pairs extracted from 5 billion webpages, with image descriptions derived from the HTML \texttt{alt-text} attribute. The version we used contains 1,666,040 unique samples.

\paragraph{CC12M:}
Following a similar collection procedure as CC3M but with more relaxed filters, CC12M consists of 12.4 million image-text pairs covering a wider range of topics and visual concepts. The version we acquired contains 7,277,079 samples.

\subsection{Downstream Datasets}
\label{appendix:downstream-datasets}

We evaluate zero-shot retrieval on Flickr30k~\cite{flickr30k} and MSCOCO~\cite{mscoco}, and classification on ImageNet~\cite{imagenet} and 10 downstream datasets. Details are provided in \cref{tab:downstream}.

\begin{table}[!htbp]
\centering
\caption{Details of downstream classification datasets.}
\label{tab:downstream}
\small
\begin{tabular}{lccccc}
\toprule
Dataset & Metric & Categories & Train Size & Test Size \\
\midrule
Food-101~\cite{food101} & Accuracy & 101 & 75,750 & 25,250 \\
CIFAR-10~\cite{cifar10} & Accuracy & 10 & 50,000 & 10,000 \\
CIFAR-100~\cite{cifar10} & Accuracy & 100 & 50,000 & 10,000 \\
SUN397~\cite{sun} & Accuracy & 397 & 19,850 & 19,850 \\
Stanford Cars~\cite{stanfordcars} & Accuracy & 196 & 8,144 & 8,041 \\
FGVC Aircraft~\cite{aircraft} & Mean per class & 100 & 6,667 & 3,333 \\
DTD~\cite{dtd} & Accuracy & 47 & 3,760 & 1,880 \\
Oxford Pets~\cite{oxfordpets} & Mean per class & 37 & 3,680 & 3,669 \\
Caltech-101~\cite{caltech101} & Mean per class & 102 & 3,060 & 6,085 \\
Oxford Flowers~\cite{flowers102} & Mean per class & 102 & 2,040 & 6,149 \\
ImageNet~\cite{imagenet} & Accuracy & 1,000 & 1,281,167 & 50,000 \\
Flickr30k~\cite{flickr30k} & Recall@$k$ & -- & 29,000 & 1,000 \\
MSCOCO~\cite{mscoco} & Recall@$k$ & -- & 113,287 & 5,000 \\
\bottomrule
\end{tabular}
\end{table}


\section{Implementation Details}
\label{appendix:implementation-details}

We build on the OpenCLIP~\cite{open_clip} codebase and evaluate using CLIP Benchmark~\cite{clip_benchmark}.

\subsection{Encoders}
\label{appendix:encoders}

We use ViT-B/16, ViT-B/32~\cite{dosovitskiy2021vit}, and ResNet-50 (RN-50) as vision encoders. All models are pretrained on CC3M, while ViT-B/16 and ViT-B/32 are additionally trained on CC12M. Following CLIP~\cite{radford2021clip}, the text encoder is a 12-layer Transformer with vocabulary size 49,408 and maximum context length 77.

\subsection{Hyperparameters}
\label{appendix:hyperparameters}

We use the same training configuration across all architectures and both pretraining datasets. The only difference between architectures is the image encoder itself; all other hyperparameters, including those for synthetic negative generation, are kept identical. Detailed configurations are provided in \cref{tab:hyperparams_training,tab:hyperparams_snap}.

\begin{table}[!htbp]
\centering
\caption{\textbf{Hyperparameters.} \textbf{(a)} CLIP/FLIP training configuration, identical for ViT-B/16, ViT-B/32, and RN-50 on both CC3M and CC12M. \textbf{(b)} \model synthetic negative generation parameters.}
\label{tab:hyperparams}
\scriptsize
\setlength{\tabcolsep}{2.6pt}
\begin{subtable}[t]{0.48\textwidth}
\centering
\caption{CLIP/FLIP hyperparameters.}
\label{tab:hyperparams_training}
\begin{tabular}{lc}
\toprule
Config & Value \\
\midrule
Image resolution & $224 \times 224$ \\
Max text length & 77 \\
Embedding dimension $d$ & 512 \\
Batch size $|\mathcal{B}|$ & 4,096 \\
Optimizer & AdamW~\cite{adamw} \\
Learning rate & $5 \times 10^{-4}$ \\
Weight decay & 0.2 \\
Adam $(\beta_1, \beta_2)$ & $(0.9, 0.98)$ \\
Adam $\epsilon$ & $1 \times 10^{-8}$ \\
LR schedule & Cosine decay \\
Warmup steps & 10,000 \\
Total epochs & 32 \\
Temperature $\tau$ (init) & 0.07 (learnable) \\
GPUs & 4$\times$ NVIDIA A100 \\
\bottomrule
\end{tabular}
\end{subtable}
\hfill
\begin{subtable}[t]{0.48\textwidth}
\centering
\caption{\model hyperparameters.}
\label{tab:hyperparams_snap}
\begin{tabular}{lc}
\toprule
Config & Value \\
\midrule
Top-$N$ hard negatives & 256 \\
Negatives per strategy & 32 \\
Number of strategies & 2 \\
Total syn.\ negatives / query & 64 \\
Temperature $\tau$ & 0.07 (constant) \\
\midrule
\multicolumn{2}{l}{\textit{Strategy hyperparameters}} \\
\midrule
Interpolation $\alpha_k$ & $\sim \mathcal{U}(0, 0.5)$ \\
Extrapolation $\beta_k$ & $\sim \mathcal{U}(1, 1.5)$ \\
Mixup $\gamma_k$ & $\sim \mathcal{U}(0, 1)$ \\
Noise std $\sigma$ & 0.01 \\
Gradient step $\delta$ & 0.01 \\
Adversarial $\eta$ & 0.01 \\
\bottomrule
\end{tabular}
\end{subtable}
\end{table}

\subsubsection{Distributed mining and gradient flow.}
Each GPU selects hard negatives for its local queries from the
globally gathered current batch, excluding the matched positives.
Gathering is performed without gradients, and source embeddings
are detached during synthesis. Gradients from synthetic-negative
terms flow through the query embeddings.


\section{Related Work}

\begin{figure}[!h]
    \centering
    \includegraphics[width=0.95\linewidth]{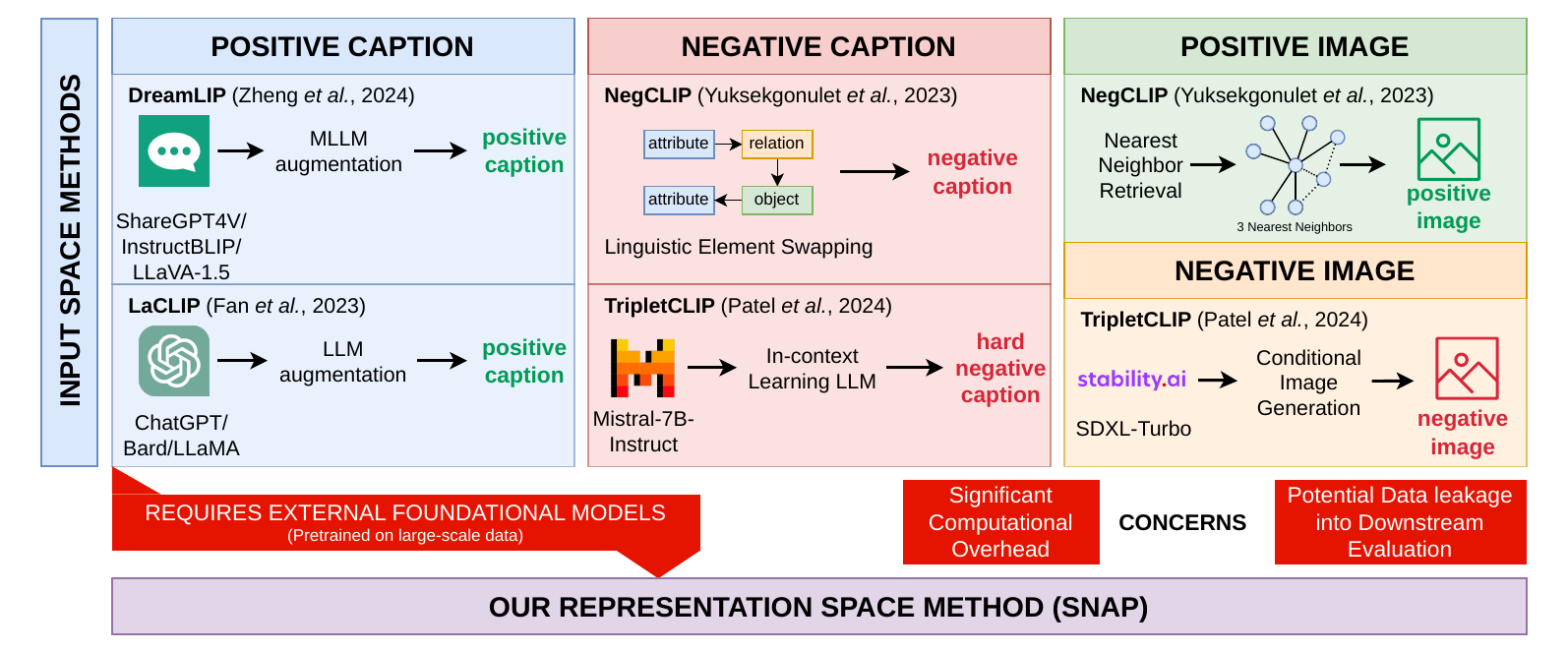}
    \caption{\textbf{Comparison of vision-language pretraining methods using synthetic data.} 
    Top: Existing \textbf{input-space methods} rely on external foundation models (e.g., LLMs for text, Diffusion for images) to generate synthetic data, incurring significant computational overhead and risk of data leakage. 
    Bottom: \textbf{\model} operates directly in the \textbf{representation space}, using in-batch embeddings to symmetrically generate hard negatives for both modalities.}
    \label{fig:sota}
\end{figure}

We briefly contrast \model with the most closely related vision-language methods that incorporate synthetic data during pretraining (\cf \cref{fig:sota}). NegCLIP~\cite{yuksekgonul2023negclip} constructs hard negative captions by swapping linguistic elements (attributes, relations, objects) in the original text and retrieves positive images from the three nearest neighbors in the training set. LaCLIP~\cite{fan2023laclip} augments the text side with synthetic positive captions generated by large language models such as ChatGPT, Bard, and LLaMA. DreamLIP~\cite{zheng2024dreamlip} follows a similar philosophy but leverages multimodal LLMs (ShareGPT4V, LLaVA, InstructBLIP) to produce richer positive captions grounded in visual content. TripletCLIP~\cite{patel2024tripletclip} operates on both modalities: it generates hard negative captions via in-context learning with Mistral-7B-Instruct-v0.2 and then synthesizes corresponding negative images using SDXL-Turbo conditioned on those captions. SPEC~\cite{peng2024synthesize} synthesizes images with controlled variations in object size, position, existence, and count to improve fine-grained vision-language understanding. All five methods operate in the input space and depend on external foundation models pretrained on large-scale data, introducing both computational overhead and potential data leakage from the generative models into downstream evaluation. In contrast, \model generates hard negatives directly in the representation space from in-batch embeddings alone, requires no external models or additional data, and applies to both modalities symmetrically with less than 10\% training time overhead. 

\section{Further Ablations and Results}
\label{sec:further-ablations}

We conduct additional ablation studies on CC3M with ViT-B/16, reporting zero-shot classification accuracy on IN-val and IN-v2.

\begin{table*}[!tbhp]
    \centering
    \begin{minipage}[t]{0.4\linewidth}
        \centering
        \caption{\textbf{Batch size ablation.}}
        \label{tab:batch_ablation}
        \begin{tabular}{lccc}
            \toprule
            \bf Batch Size & \bf Method & \bf IN-val & \bf IN-v2 \\
            \midrule
            \multirow{2}{*}{1024} & CLIP & 9.2 & 7.2 \\
            & + \model & 9.5 & 8.3 \\
            \midrule
            \multirow{2}{*}{2048} & CLIP & 9.8 & 7.8 \\
            & + \model & 10.1 & 8.9 \\
            \midrule
            \multirow{2}{*}{4096} & CLIP & 10.4 & 8.4 \\
            & + \model & \textbf{11.0} & \textbf{9.6} \\
            \bottomrule
        \end{tabular}
    \end{minipage}
    \hfill
    \begin{minipage}[t]{0.52\linewidth}
        \centering
        \caption{\textbf{Synthetic negatives ablation.}}
        \label{tab:k_ablation}
        \begin{tabular}{cccc}
            \toprule
            \bf $|\mathcal{S}|$ & \bf $|\mathcal{S}_{\text{txt}}|$ & \bf IN-val & \bf IN-v2 \\
            \midrule
            32 & 16 & 10.4 & 8.4 \\
            64 & 32 & 10.5 & 8.5 \\
            \cellcolor{gray!10}128 & \cellcolor{gray!10}64 & \textbf{11.0} & \textbf{9.6} \\
            256 & 128 & 10.7 & 8.9 \\
            512 & 256 & 9.1 & 7.0 \\
            1024 & 512 & 4.3 & 3.2 \\
            2048 & 1024 & 3.1 & 2.0 \\
            \bottomrule
        \end{tabular}
    \end{minipage}
\end{table*}

\subsection{Batch Size}

We evaluate \model at batch sizes of 1024, 2048, and 4096
(\cref{tab:batch_ablation}). \model outperforms CLIP at each tested
batch size, although its performance decline as batch size decreases
is slightly larger. These results demonstrate benefits across the
tested batch sizes. The hard-negative pool remains fixed at 256 candidates,
so its proportion of the available negatives increases as batch
size decreases(\cf \cref{tab:batch_ablation}).


\subsection{Number of Synthetic Negatives}

We vary the total number of synthetic negatives $|\mathcal{S}|=|\mathcal{S}_{\text{txt}}|+|\mathcal{S}_{\text{img}}|$, split equally between $s=3,4$. As shown in \cref{tab:k_ablation}, a small number of negatives ($|\mathcal{S}| \le 64$) yields performance similar to the CLIP baseline, as the contrastive signal is not sufficiently challenging. While $|\mathcal{S}| \ge 1024$ significantly degrades performance. At a batch size of 4096, using 512 or 1024 synthetic negatives—representing 12.5\% and 25.0\% of the real negative count, respectively—makes the contrastive task too difficult to solve.

\subsubsection{Synthesis controls.}
All synthesis controls use fixed $\tau=0.07$.
The single-strategy variants each generate 64 synthetic negatives
per query and direction, matching the total count of full \model.
The hard-negative weighting control follows the importance-sampling
approach of DiHT~\cite{radenovic2023diht}.

\subsection{Paired Seed-wise Improvements}
\label{sec:paired-seeds}

\cref{tab:paired-seeds} reports paired seed-wise improvements
for the principal CC3M ImageNet comparisons. \model improves
accuracy for every evaluated seed across all three backbones
and both baselines, showing that the average gains are
consistent across the three runs.

\begin{table}[t]
    \centering
    \small
    \setlength{\tabcolsep}{4pt}
    \caption{\textbf{Paired seed-wise improvements on CC3M.}
    ImageNet zero-shot accuracy summaries, with standard deviations shown as subscripts.
    Each paired gain $\Delta$ is the accuracy with \model minus
    the baseline accuracy for the same seed, in percentage points.
    The final column reports the mean and sample standard
    deviation of the three paired gains.}
    \label{tab:paired-seeds}
    \begin{tabular}{@{}llcccccc@{}}
        \toprule
        & & \multicolumn{2}{c}{Accuracy}
        & \multicolumn{4}{c}{Paired gain $\Delta$ (pp)} \\
        \cmidrule(lr){3-4}
        \cmidrule(l){5-8}
        Backbone & Baseline
        & Baseline & +\model
        & Seed 1 & Seed 2 & Seed 3 & Mean $\pm$ std. \\
        \midrule
        ViT-B/32 & CLIP
        & 8.0\std{3} & 8.2\std{2}
        & $+0.46$ & $+0.02$ & $+0.06$ & $+0.18 \pm 0.24$ \\
                 & FLIP
        & 6.8\std{3} & 7.2\std{1}
        & $+0.10$ & $+0.27$ & $+0.74$ & $+0.37 \pm 0.33$ \\
        \midrule
        ViT-B/16 & CLIP
        & 10.4\std{2} & 11.0\std{2}
        & $+1.07$ & $+0.30$ & $+0.37$ & $+0.58 \pm 0.43$ \\
                 & FLIP
        & 10.3\std{3} & 11.0\std{2}
        & $+0.88$ & $+1.16$ & $+0.12$ & $+0.72 \pm 0.54$ \\
        \midrule
        RN-50 & CLIP
        & 13.6\std{1} & 14.7\std{2}
        & $+0.99$ & $+1.05$ & $+1.32$ & $+1.12 \pm 0.18$ \\
              & FLIP
        & 13.7\std{1} & 14.9\std{1}
        & $+1.17$ & $+1.25$ & $+1.15$ & $+1.19 \pm 0.05$ \\
        \bottomrule
    \end{tabular}
\end{table}

\subsection{Compositional Evaluation}
\label{sec:compositional}

We evaluate CLIP and \model with ViT-B/16 pretrained on CC12M
on ARO~\cite{yuksekgonul2023negclip} and Winoground~\cite{thrush2022winoground} (\cref{tab:compositional}).
\model improves ARO relation and attribute accuracy by 3.6 and
4.2 percentage points, respectively, while Winoground scores
remain similar.

\begin{table}[t]
    \centering
    \caption{\textbf{Compositional evaluation.}
    Scores (\%) for ViT-B/16 pretrained on CC12M.}
    \label{tab:compositional}
    \small
    \begin{tabular}{@{}lccccc@{}}
        \toprule
        & \multicolumn{2}{c}{ARO}
        & \multicolumn{3}{c}{Winoground} \\
        \cmidrule(lr){2-3}\cmidrule(lr){4-6}
        Method & Relation & Attribute & Text & Image & Group \\
        \midrule
        CLIP & 51.2 & 52.1 & 24.5 & \textbf{9.8} & 6.3 \\
        + \model & \textbf{54.8} & \textbf{56.3}
                 & \textbf{25.2} & 9.6 & \textbf{6.5} \\
        \bottomrule
    \end{tabular}
\end{table}


\section{Future Work}

An immediate direction is to study SNAP at larger scales: training on web-scale datasets, exploring deeper and wider architectures, and integrating with recent advances in text and image encoders. The top-similarity hard-negative pool may also contain semantically compatible
but unmatched examples, particularly in web-crawled data. \model does not explicitly
filter such false negatives; incorporating semantic filtering or auditing their prevalence
is left for future work.


\section{Limitations} 

Our experiments are conducted on CC3M and CC12M, which are relatively small-scale compared to proprietary datasets used by methods such as CLIP (400M pairs) and ALIGN (1B pairs). Training on larger datasets was not feasible given our computational budget of 4$\times$ A100 GPUs, where a single CC12M run already requires multiple days. Scaling to larger data and model sizes, while remaining environmentally conscious, is an important direction for future work. We observe broad improvements across the evaluated settings; whether
these benefits persist at larger data and model scales remains to be tested.


\section{Discussion} 

\model operates as a drop-in augmentation to any contrastive framework that uses an InfoNCE-style loss with a softmax denominator over negatives. Since synthetic negatives are generated from in-batch embeddings alone, the method is agnostic to the choice of encoders, data, and training procedure. We expect \model to be applicable beyond the CLIP~\cite{radford2021clip} and FLIP~\cite{li2023flip} frameworks evaluated here, including methods such as SLIP~\cite{mu2022slip}, DeCLIP~\cite{li2022declip}, FLIP~\cite{li2023flip}, LiT~\cite{zhai2022lit}, LaCLIP~\cite{fan2023laclip}, DreamLIP~\cite{zheng2024dreamlip}, CyCLIP~\cite{goel2022cyclip}, and TripletCLIP~\cite{patel2024tripletclip}, all of which build on the same InfoNCE-style contrastive objective.


\section{Reproducibility}
\label{appendix:reproducibility}

We build on the OpenCLIP~\cite{open_clip} codebase for all training and use CLIP Benchmark~\cite{clip_benchmark} for all downstream evaluations. Pretraining datasets were downloaded using img2dataset~\cite{img2dataset}. All experiments use identical hyperparameters across architectures and datasets (\cf \cref{tab:hyperparams}). Training was performed on 4$\times$ NVIDIA A100 80GB GPUs using PyTorch with mixed-precision (FP16) training. We use the same set of random seeds across all runs for reproducibility. \model requires no external models, no additional data, and no offline preprocessing---the only modification to the standard OpenCLIP training loop is the synthetic negative generation module, which operates entirely in the representation space during the forward pass as detailed in \cref{alg:snap,alg:strategies}. We will release our code and pretrained model weights upon acceptance to facilitate reproduction of all reported results.


\section{Algorithm}
\label{appendix:algorithm}

\Cref{alg:snap} provides the pseudo-code of \model, followed by the detailed implementation of the two specific generation strategies, $s=3$ and $s=4$, used in our approach (\Cref{alg:strategies}).

\begin{algorithm*}[!p]
\caption{Pseudocode of \model in a PyTorch-like style.}
\label{alg:snap}
\algcomment{\fontsize{7.2pt}{0em}\selectfont 
\texttt{mm}: matrix multiplication; 
\texttt{cat}: concatenation; 
\texttt{topk}: returns indices of top-k values;
\texttt{einsum}: Einstein summation;
\texttt{normalize}: $\ell_2$ normalization along dimension.
}\definecolor{codeblue}{rgb}{0.25,0.5,0.5}
\lstset{
  backgroundcolor=\color{white},
  basicstyle=\fontsize{7.2pt}{7.2pt}\ttfamily\selectfont,
  columns=fullflexible,
  breaklines=true,
  captionpos=b,
  commentstyle=\fontsize{7.2pt}{7.2pt}\color{codeblue},
  keywordstyle=\fontsize{7.2pt}{7.2pt},
}
\begin{lstlisting}[language=python]
# f_v, f_t: visual and text encoders
# t: fixed temperature parameter
# n_hard: pool size for sampling anchors (e.g., 256)
# s3, s4: number of synthetic negatives per strategy (default: 32 each)

# t: fixed temperature
# n_hard: hard-negative pool size
# s3, s4: synthetic negatives per strategy and direction

def synthetic_logits(query, candidates, labels):
    scores = query @ candidates.T
    scores = scores.clone()
    scores[arange(query.shape[0]), labels] = -inf
    idxs = topk(scores, k=n_hard, dim=1).indices

    mix = generate_s3(candidates, idxs, s3)
    noise = generate_s4(candidates, idxs, s4)
    synthetic = cat([mix, noise], dim=1)
    return einsum("nc,nkc->nk", query, synthetic)

for images, texts in loader:
    v = normalize(f_v(images), dim=-1)
    u = normalize(f_t(texts), dim=-1)

    # Global current-batch candidates, detached for synthesis
    all_v = gather_no_grad(v.detach())
    all_u = gather_no_grad(u.detach())

    # Positive indices in the globally gathered batch
    # Assumes equal local batch sizes and rank-ordered gathering
    labels = rank * v.shape[0] + arange(v.shape[0])

    syn_i2t = synthetic_logits(v, all_u, labels)
    syn_t2i = synthetic_logits(u, all_v, labels)

    # Obtain base logits with the existing CLIP gradient handling
    base_i2t, base_t2i = clip_logits(v, u)

    logits_i2t = cat([base_i2t, syn_i2t], dim=1) / t
    logits_t2i = cat([base_t2i, syn_t2i], dim=1) / t

    loss = 0.5 * (
        cross_entropy(logits_i2t, labels)
        + cross_entropy(logits_t2i, labels)
        )

    optimizer.zero_grad()
    loss.backward()
    optimizer.step()
\end{lstlisting}
\hrule
\end{algorithm*}

\begin{algorithm*}[!p]
\caption{\model Generation Strategies in a PyTorch-like style.}
\label{alg:strategies}
\definecolor{codeblue}{rgb}{0.25,0.5,0.5}
\lstset{
  backgroundcolor=\color{white},
  basicstyle=\fontsize{7.2pt}{7.2pt}\ttfamily\selectfont,
  columns=fullflexible,
  breaklines=true,
  captionpos=b,
  commentstyle=\fontsize{7.2pt}{7.2pt}\color{codeblue},
  keywordstyle=\fontsize{7.2pt}{7.2pt},
}
\begin{lstlisting}[language=python]
def generate_s3(t_feat, idxs_hard, num_neg, gamma=1.0):
    """Mixup between two hard negatives."""
    # sample two sets of indices from the hard pool
    idx1 = randint(0, n_hard, size=(batch_size, num_neg))
    idx2 = randint(0, n_hard, size=(batch_size, num_neg))
    
    h1 = t_feat[gather(idxs_hard, 1, idx1)]
    h2 = t_feat[gather(idxs_hard, 1, idx2)]
    
    lam = rand(batch_size, num_neg, 1) * gamma
    s3 = lam * h1 + (1 - lam) * h2
    return normalize(s3, dim=-1)

def generate_s4(t_feat, idxs_hard, num_neg, sigma=0.01):
    """Noise injection into a hard negative."""
    idx = randint(0, n_hard, size=(batch_size, num_neg))
    h = t_feat[gather(idxs_hard, 1, idx)]
    
    noise = randn_like(h) * sigma
    return normalize(h + noise, dim=-1)
\end{lstlisting}
\hrule
\end{algorithm*}

\end{document}